\documentclass[pdflatex,sn-mathphys-num]{sn-jnl}

\usepackage{graphicx}%
\usepackage{multirow}%
\usepackage{amsmath,amssymb,amsfonts}%
\usepackage{amsthm}%
\usepackage{mathrsfs}%
\usepackage[title]{appendix}%
\usepackage{xcolor}%
\usepackage{textcomp}%
\usepackage{manyfoot}%
\usepackage{booktabs}%
\usepackage{algorithm}%
\usepackage{algorithmicx}%
\usepackage{algpseudocode}%
\usepackage{listings}%
\usepackage{tabularx}%
\usepackage{array}     
\usepackage{float}     
\usepackage{rotating}
\usepackage{acronym}

\acrodef{ai}[AI]{Artificial Intelligence}
\acrodef{xai}[XAI]{eXplainable Artificial Intelligence}
\acrodef{ml}[ML]{Machine Learning}
\acrodef{dl}[DL]{Deep Learning}
\acrodef{cnn}[CNN]{Convolutional Neural Network}
\acrodef{svm}[SVM]{Support Vector Machine}
\acrodef{gradcam}[Grad-CAM]{Gradient-weighted Class Activation Mapping}
\acrodef{shap}[SHAP]{SHapley Additive exPlanations}

\newcommand{\enquote}[1]{``#1''}

\theoremstyle{thmstyleone}%
\theoremstyle{thmstyletwo}%

\theoremstyle{thmstylethree}%

\begin{document}

\title[Article Title]{Synthetic Data Augmentation for Multi-Task Chinese Porcelain Classification: A Stable Diffusion Approach}


\author[1]{\fnm{Ziyao} \sur{Ling}}\email{ziyao.ling2@unibo.it}

\author[1]{\fnm{Silvia} \sur{Mirri}}\email{silvia.mirri@unibo.it}

\author[1]{\fnm{Paola} \sur{Salomoni}}\email{paola.salomoni@unibo.it}

\author*[1]{\fnm{Giovanni} \sur{Delnevo}}\email{giovanni.delnevo2@unibo.it}

\affil*[1]{\orgdiv{Department of Computer Science and Engineering}, \orgname{University of Bologna}, \orgaddress{\street{Mura Anteo Zamboni 7}, \city{Bologna}, \postcode{40127}, \state{Emilia-Romagna}, \country{Italy}}}


\abstract{The scarcity of training data presents a fundamental challenge in applying deep learning to archaeological artifact classification, particularly for the rare types of Chinese porcelain. This study investigates whether synthetic images generated through Stable Diffusion with Low-Rank Adaptation (LoRA) can effectively augment limited real datasets for multi-task CNN-based porcelain classification. Using MobileNetV3 with transfer learning, we conducted controlled experiments comparing models trained on pure real data against those trained on mixed real-synthetic datasets (95:5 and 90:10 ratios) across four classification tasks: dynasty, glaze, kiln and type identification. Results demonstrate task-specific benefits: type classification showed the most substantial improvement (5.5\% F1-macro increase with 90:10 ratio), while dynasty and kiln tasks exhibited modest gains (3-4\%), suggesting that synthetic augmentation effectiveness depends on the alignment between generated features and task-relevant visual signatures. Our work contributes practical guidelines for deploying generative AI in archaeological research, demonstrating both the potential and limitations of synthetic data when archaeological authenticity must be balanced with data diversity.
}




\maketitle

\section{Introduction}
\label{sec:intro}

Chinese porcelain, as a significant component of cultural heritage, embodies profound historical and artistic value. Contemporary porcelain authentication methods remain predominantly dependent on expert connoisseurship, with modern scientific techniques serving merely as auxiliary tools rather than replacement methodologies \cite{wang2019brief}. Consequently, the field of porcelain authentication currently faces two primary challenges: first, the time-consuming nature of manual authentication coupled with the inherent subjectivity of expert judgment; second, the prohibitive costs associated with advanced analytical techniques, some of which, such as thermoluminescence dating may cause irreversible damage to the artifacts \cite{portakal2024assessment}.

Within this context, deep learning has demonstrated substantial potential in augmenting Chinese porcelain authentication, primarily through the deployment of Convolutional Neural Networks (CNNs) for artifact classification based on distinctive features including decorative motifs, morphological forms, and chronological periods \cite{han2023typical}. Existing approaches encompass both single-feature classification models, which focus on individual attributes, and multi-task learning frameworks that simultaneously classify multiple characteristics \cite{ling2025multi}. These methodological choices are grounded in traditional morphological analysis, which emphasizes the tripartite schema of color, form, and pattern as fundamental diagnostic criteria \cite{2008maweidu}. However, the fragility and non-renewable nature of porcelain artifacts as cultural heritage objects presents significant challenges for comprehensive data collection \cite{choy2016unesco}. Current research in this domain is constrained by limited dataset sizes, Weng et al., selected porcelain of the same shape but from five different ages, with only 10 images from each age, a total of 50 images for training \cite{weng2017machine}. Niu et al. collected 5000 images, which represents the largest porcelain dataset reported in the literature to date, yet this remains insufficient compared to successful deep learning applications in other domains, where datasets typically contain tens of thousands to hundreds of thousands of images \cite{niu2022using}. Given the high degree of intra-class variability exhibited by porcelain artifacts, where individual pieces may display unique characteristics arising from variations in production techniques, raw materials, and firing conditions, this data scarcity causes limitations, including overfitting and poor generalization to unseen porcelain samples \cite{weng2017machine}. Consequently, dataset augmentation represents one of the challenges confronting the application of deep learning to porcelain authentication, necessitating innovative approaches to expand training data while maintaining archaeological authenticity \cite{ottoni2025deep}.

    Beyond the issue of limited dataset size, current mitigation strategies using traditional augmentation techniques, including rotation, flipping, and color jittering, while effective for increasing sample counts, are fundamentally constrained by the original image content \cite{shorten2019survey}. They cannot introduce genuinely new visual variations such as alternative viewing angles, varied lighting conditions that reveal different glaze properties, or morphological variations within a porcelain type. For archaeological artifacts, where each piece represents centuries of material culture and craftsmanship, these limitations are particularly acute. Traditional augmentation merely permutes existing pixels rather than synthesizing new visual information that could capture the rich diversity within each porcelain category. For instance, a single type of Song dynasty celadon bowl might exhibit substantial variations in glaze thickness, crackle patterns, color gradients, and surface textures depending on its specific production batch, kiln position, and firing conditions that cannot be recreated through geometric or photometric transformations alone \cite{2008maweidu}. Moreover, the limited number of high-quality porcelain photograph available means that traditional augmentation repeatedly recycles the same visual features, potentially leading to overfitting on porcelain-specific characteristics rather than learning generalizable class-level patterns \cite{ying2019overview}.
    
This fundamental limitation has motivated researchers to explore generative approaches that can create entirely new, photorealistic images of archaeological porcelains \cite{altaweel2024using}. Recent advances in deep generative models, particularly Generative Adversarial Networks (GANs) \cite{goodfellow2020generative} and diffusion models \cite{ho2020denoising}, offer significant capabilities to synthesize novel visual content that maintains the authentic characteristics of historical porcelains while introducing controlled variations \cite{wang2025visual}. These image generation techniques have demonstrated success in addressing multiple challenges simultaneously: expanding limited datasets, balancing class distributions, and generating diverse intra-class variations \cite{rodriguez2022synthetic}. By learning the underlying distribution of visual features from existing specimens, generative models can produce synthetic porcelain images that exhibit plausible variations in glaze colours, decorative patterns, and morphological characteristics \cite{bao2025ai}

In this research, we investigate the application of diffusion-based synthetic image generation to enhance multi-task CNN classification of Chinese porcelain. Building upon our baseline multi-task MobileNetV3 model trained on real images, we explore whether synthetic data augmentation can improve performance across multiple tasks. Three main contributions of this paper can be highlighted: First, we present a comprehensive comparative analysis demonstrating that models trained on mixed real-synthetic datasets (95:5 and 90:10 ratios) using Stable Diffusion with LoRA achieve consistent improvements over the real-data-only baseline when evaluated on the same held-out test set of authentic porcelain images. Second, we provide task-specific performance analysis across four tasks (dynasty, glaze, and type), revealing differential improvements where type classification benefits most significantly from synthetic augmentation (up to 4\% improvement). In contrast, dynasty and kiln tasks show more modest gains, indicating that the effectiveness of synthetic data is task-dependent. Third, we demonstrate the practical viability of structured prompt engineering grounded in archaeological documentation for generating training-quality synthetic porcelain images, showing that carefully designed prompts can produce diverse yet archaeologically plausible variations that enhance model generalization without introducing distribution shift in the test phase.

The remainder of this paper is structured as follows. Section \ref{sec:relatedwork} outlines the related works while Section \ref{sec: materials} details the dataset and the experimental design, including strategies for synthetic dataset creation, various mixing configurations between real and synthetic data, and the evaluation framework used to assess impact across multiple dimensions of performance. Section \ref{sec:results} presents comprehensive results addressing each research question, analyzing synthetic image quality, impact on class imbalance, and the effects of different mixing strategies on model performance and generalization. Section \ref{sec:conclusion}  concludes with a summary of key findings, contributions to both computer vision and digital heritage fields, and directions for future research in synthetic cultural artifact generation.

\section{Related Work}
\label{sec:relatedwork}

The evolution of generative artificial intelligence has opened unprecedented opportunities for cultural heritage conservation and dissemination. The progression from Generative Adversarial Networks (GANs) \cite{goodfellow2020generative} to diffusion-based models \cite{ho2020denoising} has dramatically enhanced the capability to synthesize high-fidelity images with fine-grained control. In the cultural heritage domain, these technologies have been applied across diverse applications, from artifacts restoration \cite{baek2025rich, zhang2025generating} to virtual museum experiences \cite{kilis2025ai}.

Early applications of GANs in cultural heritage focused primarily on image restoration and completion tasks. For instance, Lin et al. \cite{lin2025embroidery} employed an improved GAN to reconstruct damaged embroidery cultural relics, while Cai et al. developed a style-transfer GAN for historical Chinese character recognition. However, these approaches often struggled with maintaining cultural authenticity, particularly when generating novel content rather than restoring existing artifacts.

The emergence of diffusion models has substantially improved generation quality and controllability for heritage applications. Stable Diffusion \cite{rombach2022high} and its variants have shown particular promise in generating culturally-specific imagery. Chen et al. \cite{chen2025ancient} successfully applied a mural images super-resolution network based on the diffusion model to generate high-fidelity mural images with rich details, while Chu et al. \cite{bi2025chu} developed a diffusion-based approach for synthesizing Chu-style lacquerware images. These successes demonstrate the potential for generating training data in domains where original artifacts are scarce or complex. However, archaeological porcelains present unique challenges not fully addressed by existing approaches. The subtle variations in glaze colour, form shapes and decorative patterns that distinguish authentic porcelain from different periods and kilns require domain-specific adaptations beyond general-purpose image generation models.

Several studies have specifically addressed porcelain generation, though most focus on contemporary artistic applications rather than archaeological accuracy. Bao et al. \cite{bao2025ai} used a Stable Diffusion platform and Low-Rank Adaptation (LoRA) technology to generate blue-and-white porcelain design, offering viable solutions for the contemporary reinvention of traditional crafts.  porcelain designs, achieving visually appealing results but without consideration for historical authenticity. More relevant to archaeological applications, Zhao et al \cite{zhao2025ai} developed a GAN-based system for generating synthetic porcelain images, demonstrating improved recognition performance when augmenting limited archaeological datasets. Their work showed that synthetic images could capture morphological variations within porcelain shapes and colours, though they noted challenges in reproducing realistic complex patterns.

The critical challenge in applying generative AI to archaeological porcelain lies in balancing generation diversity with historical plausibility. Unlike natural image domains where some deviation from reality may be acceptable or even desirable, synthetic porcelain images must respect complex constraints: documented glaze colors, period-appropriate decorative patterns, technological limitations of historical kilns, and regional stylistic shapes \cite{2008maweidu}. Recent work by Yan et al. \cite{yan2025cultural} on generating Chinese bronze vessels highlighted similar challenges, finding that models trained solely on visual data often produced anachronistic combinations of features from different periods. This underscores the need for incorporating domain expertise into the generation process, whether through carefully designed loss functions \cite{xie2025method}, structured prompts based on archaeological documentation (as proposed in this study), or hybrid approaches combining rule-based constraints with learned representations.

Evaluation of generated artifact images presents another significant challenge. Standard computer vision metrics like Fréchet Inception Distance (FID) \cite{heusel2017gans} and Inception Score (IS) \cite{salimans2016improved} may not adequately capture archaeological authenticity. Kuang et al. \cite{kuang2025preserving} proposed domain-specific metrics combining technical accuracy assessment with historical plausibility scores, though these require extensive expert annotation. For porcelain specifically, evaluation must consider multiple dimensions: morphological accuracy (shape and proportions), surface characteristics (glaze color, texture, crackle patterns), decorative authenticity (period-appropriate motifs and execution styles), and technical feasibility (whether the combination of features could have been produced with historical techniques). The absence of standardized evaluation protocols for synthetic archaeological images remains a significant gap in the literature, one that this study addresses through systematic comparison of classification performance on authentic test sets.

\section{Materials and Methods}
\label{sec: materials}
This Section details the methodology employed in our research. We outline the specific research questions addressed, design a comprehensive framework of the synthetic dataset generation strategies through LoRA fine-tuning model, and provide a comparative classification analysis between synthetic and traditional augmentation approaches deployed on the same multi-task CNN model(MobileNetV3). Furthermore, we define the image quality metrics, training parameters and evaluation metrics used to assess performance, and conclude with a thorough explanation of the experimental settings.

\subsection{Research Questions}
\label{subsec:RQ}
This paper investigates the potential of synthetic image generation to enhance CNN-based porcelain classification through three interconnected research questions:

\textbf{RQ1:} \textit{How does the use of synthetic porcelain images affect the performance of multi-task CNN(MobileNetV3) model?}  

\textbf{RQ1.1:} \textit{How does the quality and diversity of generated porcelain images affect classification performance when used as training data?}  

\textbf{RQ1.2:} \textit{To what extent can synthetic images reduce label imbalance in real datasets, and what is their impact on model performance for underrepresented classes?}  

\textbf{RQ1.3:} \textit{How does mixing real and synthetic images in different proportions affect multi-task CNN(MobileNetV3) classifier stability, overfitting, and generalization?}

\subsection{Dataset Description}

\label{subsec: dataset}

To evaluate the impact of synthetic augmentation on CNN classification, we constructed three training dataset configurations. The \textbf{Baseline (Real-only)} set contains 25,877 real porcelain images, exhibiting severe class imbalance with nine glaze-type combinations having only five samples each. The \textbf{Real + Synthetic-570} set introduces 570 synthetic images (5\% increase) targeting extremely rare categories and confusion pairs, while the \textbf{Real + Synthetic-2500} set adds 2500 synthetic images (10\% increase) for broader coverage.

\subsubsection{Real Images}
\label{subsubsec: realimages}
The Song and Yuan dynasties (960-1368 CE) represent the golden age of Chinese porcelain achievement, witnessing the presence of two major production systems during the Song period: the imperial kiln system, comprising the Five Great Kilns (Ru, Guan, Ge, Jun, and Ding), and the folk kiln system, including the Cizhou, Yaozhou, Longquan, and Jingdezhen kilns \cite{2008maweidu}. Thus, this period constitutes the first peak in Chinese porcelain history. Due to the diversity and complex system of archaeological Chinese porcelains, data collection requires multiple sources to cover dynasties, kilns, glazes and object types. This study focuses on the porcelain dynasties from the Song (960–1279 CE) and Yuan (1271–1368 CE) dynasties.
Sample images of porcelains are collected from the open data of the National Palace Museum in Taipei and the Palace Museum in Beijing, as described in Table \ref{tab:sources}.

\begin{table}[htbp]
  \centering
  \caption{Image Sources and Licences for Museum Digital Collections}
  \begin{tabularx}{\textwidth}{@{} l l X @{}}
    \toprule
    Code & Museum & Licence \\ \midrule
    PMBJ & Palace Museum, Beijing
         & Images obtained through \url{https://www.dpm.org.cn/explore/collections} \\[2pt]
    PMTP & National Palace Museum in Taipei
         & CC-BY-4.0; images obtained through \url{https://digitalarchive.npm.gov.tw/opendata} \\
    \bottomrule
  \end{tabularx}
  \label{tab:sources}
\end{table}

After eliminating the images that did not meet the experimental requirements, we retained 7,263 high-resolution porcelain images from the Song and Yuan dynasties, reported in Table \ref{tab:totals}. Each image was captured under controlled lighting conditions with standardised colour calibration to ensure consistency across different sources. This dataset includes different glazes, colours, kilns and object types. Period attributions and related metadata were provided by archaeologists affiliated with the contributing museums and are publicly available on their websites, ensuring both authenticity and diversity of sources.
Finally, culturally sensitive artefacts were masked before publication, and all licences were verified to permit non-commercial academic use. 

\begin{table}[htbp]
  \centering
  \caption{Distribution of Song and Yuan dynasty porcelain images by data source}
  \begin{tabular}{p{3.5cm}p{2.5cm}p{2.5cm}p{2cm}}
    \toprule
    Source & Song & Yuan & Total \\ \midrule
    PMBJ   &   358 &   91 &   449 \\
    PMTP   & 5\,304 & 1\,510 & 6\,814 \\ \midrule
    \textbf{Total} & \textbf{5\,662} & \textbf{1\,601} & \textbf{7\,263} \\
    \bottomrule
  \end{tabular}
  \label{tab:totals}
\end{table}

\begin{figure}[htbp]
    \centering
    \includegraphics[width=0.9\textwidth]{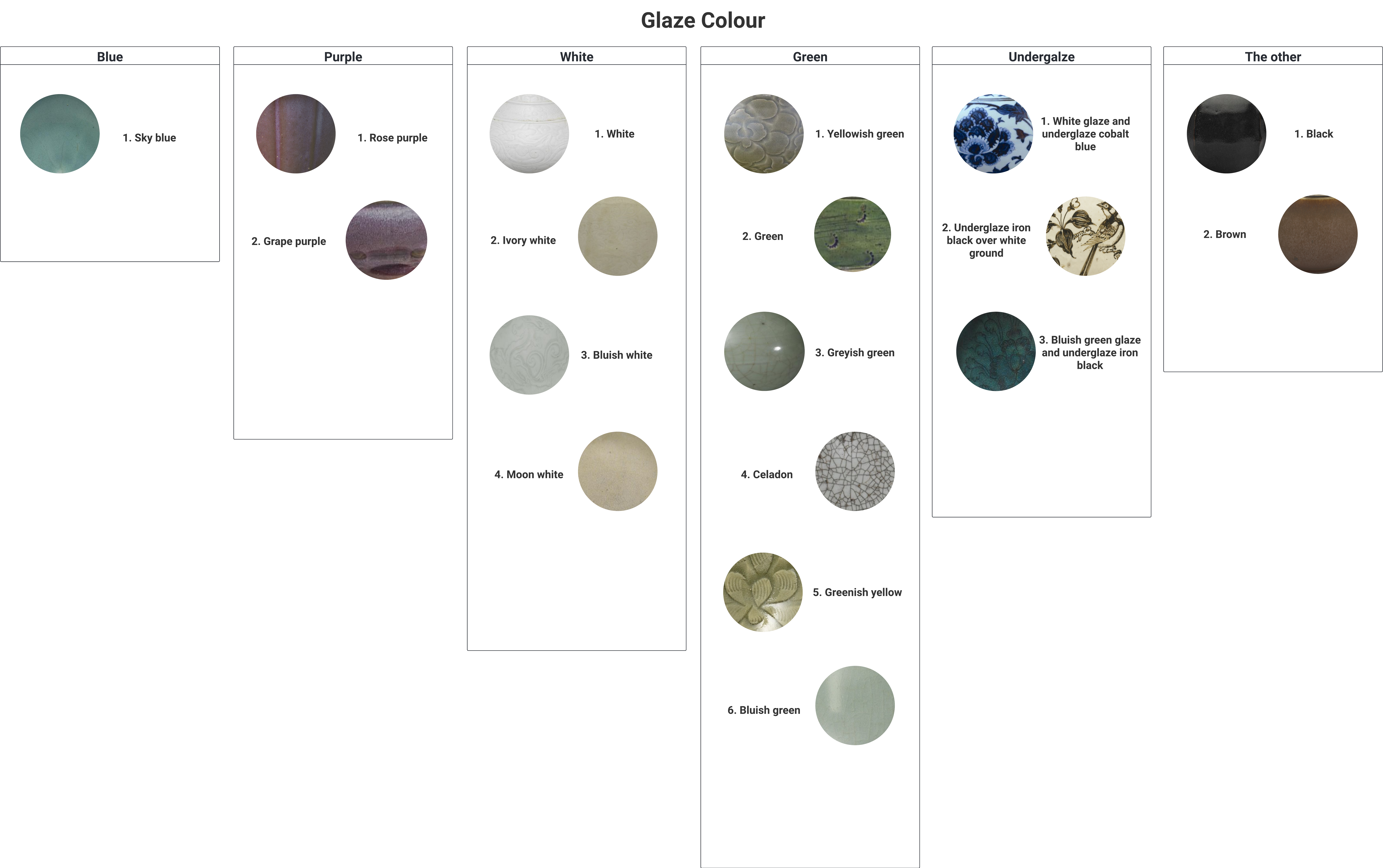}
     \begin{tablenotes}[para,flushleft] 
      \footnotesize
      \textit{Note.} All images are drawn from the dataset compiled for this study.
    \end{tablenotes}
    \caption{The Chinese porcelain glaze colour categories from the dataset}\label{fig:glazecolour}
\end{figure}

The traditional morphological authentication of porcelains is based on three key elements: colour, form(shape) and pattern \cite{2010maoxiaolu}. Porcelain surfaces are typically coated with one or two layers of glaze. The majority of glazes exhibit colouration, with even white glazes and transparent glazes demonstrating some degree of colouration, such as celadon or ivory white. Pure, colourless, transparent glazes are not present in archaeological porcelains. During the Song and Yuan dynasties, black glaze, blue glaze, brown glaze, and purple glaze were all produced using iron oxide as the colouring agent, with variations occurring only in the proportion of iron content and the firing conditions \cite{2010maoxiaolu}, as presented in Figure \ref{fig:glazecolour}. 

Its practical application and aesthetic appeal principally dictate the type of porcelain. Furthermore, it evolves by temporal shifts and advancements in porcelain production techniques. The typological classification of porcelain forms presented in this dataset categorises porcelain objects into major functional groups such as washers, pots, cups, boxes, pillows, vases, bowls, plates, planters, incense-burners, jars, and zuns, each further subdivided into specific shapes, as described in Figure \ref{fig:formclass}.

\begin{figure}[htbp]
    \centering
    \includegraphics[width=0.9\textwidth]{images/formclass.pdf}
     \begin{tablenotes}[para,flushleft] 
      \footnotesize
      \textit{Note.} All images are drawn from the dataset compiled for this study.
    \end{tablenotes}
    \caption{The Chinese porcelain form categories from the dataset}\label{fig:formclass}
\end{figure}

\begin{figure}[htbp]
    \centering
    \includegraphics[width=0.9\textwidth]{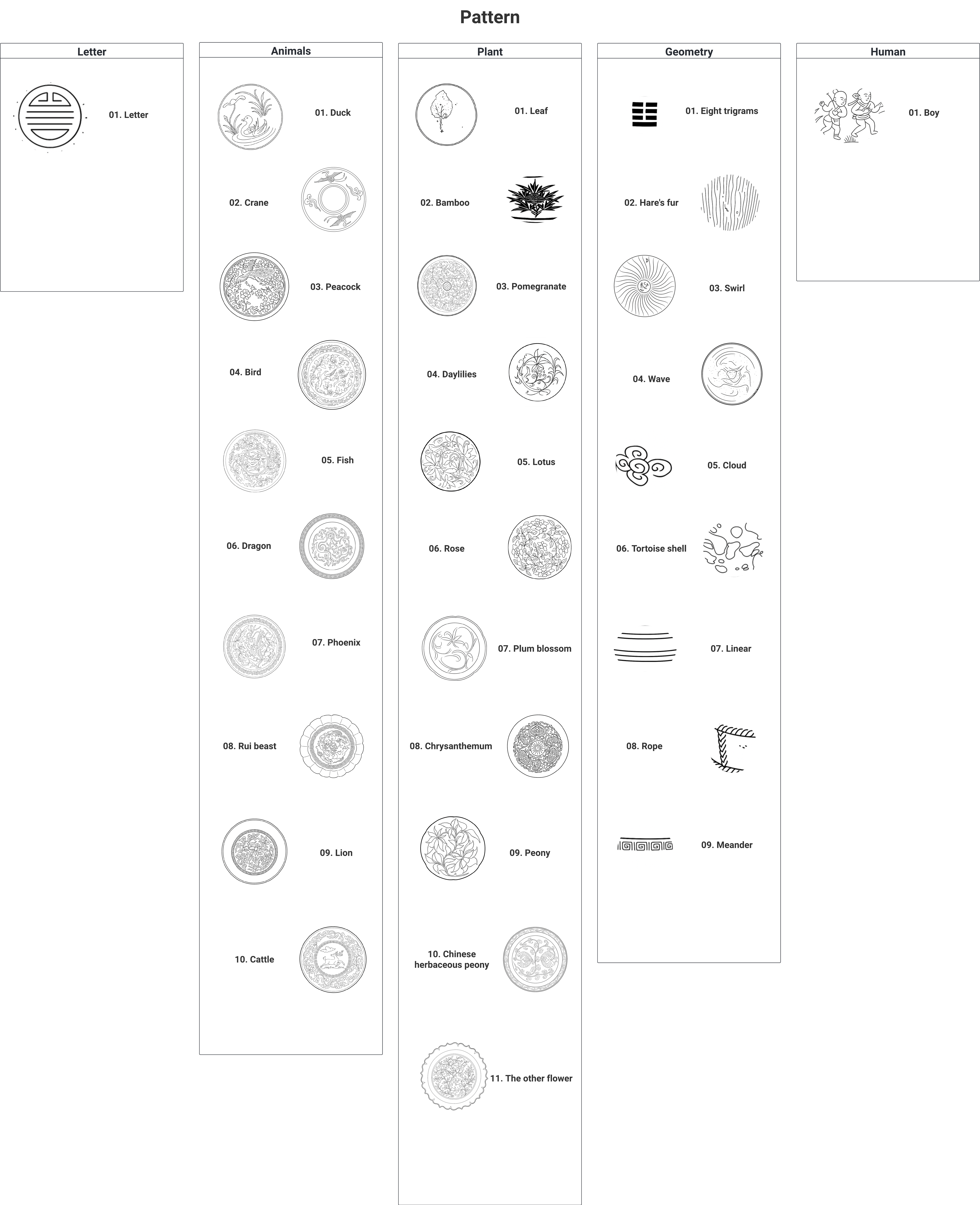}
     \begin{tablenotes}[para,flushleft] 
      \footnotesize
      \textit{Note.} All images are drawn from the dataset compiled for this study.
    \end{tablenotes}
    \caption{The Chinese porcelain pattern categories from the dataset}\label{fig:formclass2}
\end{figure}

Decorative patterns constitute a vital aspect of porcelain morphology, particularly from the Yuan dynasty onwards. The classification of patterns reflects four production techniques, such as printed, incised, carved and painted. As shown in Figure \ref{fig:formclass2}, this figure presents a systematic taxonomy of decoration patterns found on Chinese porcelains, organised into five major categories, such as letter, animals, plant, geometry and human, each further subdivided into specific patterns. For instance, plants include leaves, bamboo, daylilies, lotus and rose. 

Statistical analysis of our dataset confirms this historical reality: only 24\% (n=1,746) of the 7,263 porcelain images contain identifiable decorative patterns, with the remaining 76\% consisting of undecorated monochrome porcelains. Among the decorated subset, the distribution across the 31 pattern categories exhibits extreme imbalance. This severe class imbalance, compounded by the overall scarcity of decorated porcelains, presented significant methodological challenges that ultimately led to the exclusion of pattern classification from our multi-task framework. The decision to omit pattern recognition as a fifth task was further justified by technical considerations. 

\begin{figure}[htbp]
    \centering
    \includegraphics[width=1.1\textwidth]{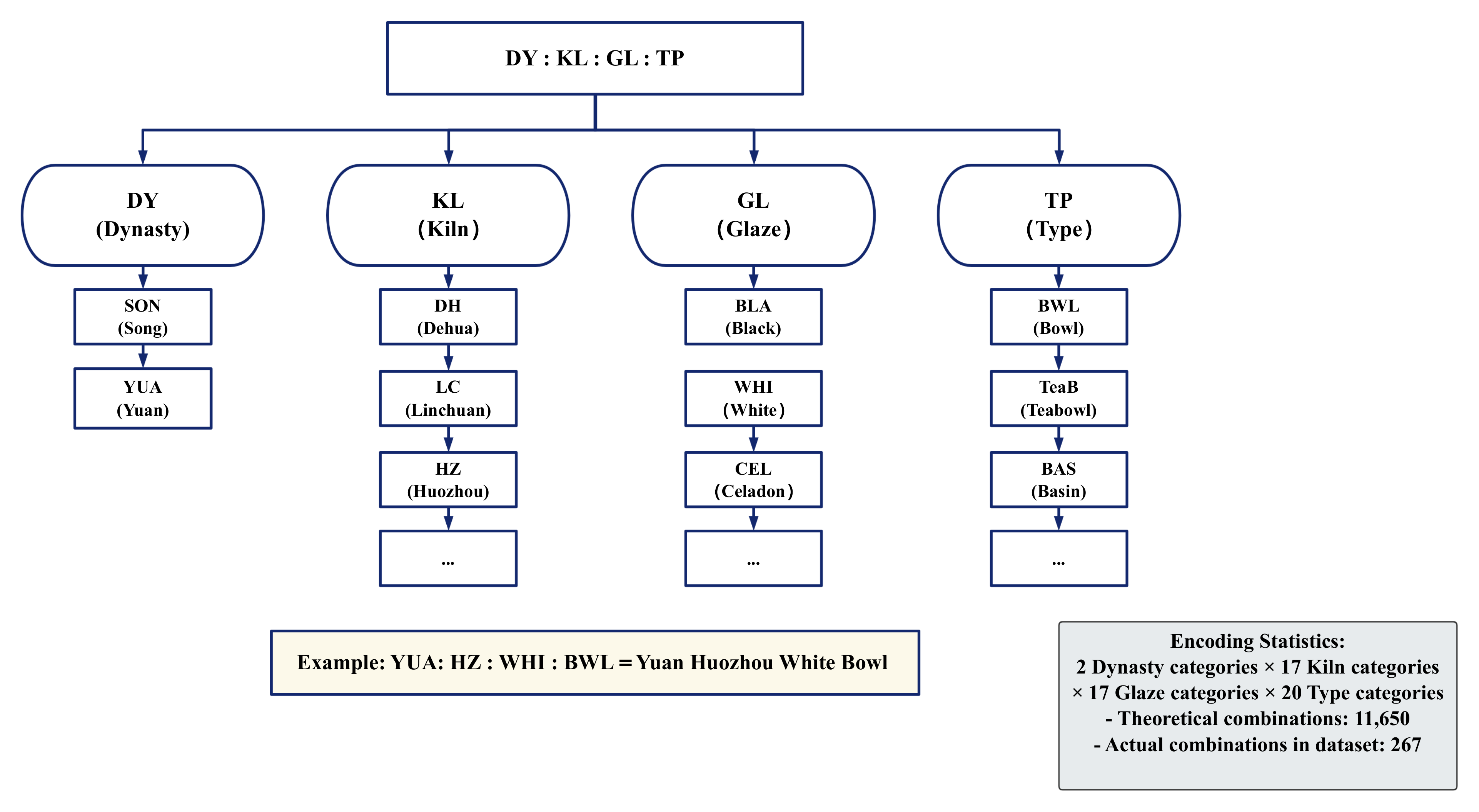}
    \begin{tablenotes}[para,flushleft] 
      \footnotesize
      \textit{Note.} Each attribute uses a 2-3 letter mnemonic code for dataset operations and unambiguous identification.
    \end{tablenotes}
    \caption{Hierarchical Label System for Porcelain Classification} \label{fig:hierarchical}
\end{figure}

According to the traditional morphological authentication of porcelain, the image dataset is annotated with a four-axis hierarchical code, where DY denotes the political period (Dynasty), KL(kiln) denotes the producing kiln, GL (Glaze) denotes the glaze coulors, and TP (Type) denotes the functional porcelain type. The hierarchical label system shown in Figure \ref{fig:hierarchical} encodes each porcelain artifact using four attributes (Dynasty, Kiln, Glaze, Type), resulting in 267 observed combinations out of 11,650 theoretical possibilities. This sparse representation (2.3\%) reflects the historical and technological constraints of porcelain production. The code performs a dual function: it provides an unambiguous primary key for dataset operations, and it embeds the principal variables that condition stylistic and technological change in Chinese porcelains. Each attribute is represented by a stable two- or three-letter mnemonic, the annotation process utilised the Labelbee tool, labelling images with four corresponding tags and saving them in CSV format. 

We implemented an adaptive splitting algorithm that adjusts the partitioning strategy based on combination frequency. For singleton combinations (n=1), samples were assigned exclusively to the training set to maximise learning opportunity for these extremely rare cases. Doublet combinations (n=2) were deterministically split between validation and test sets, ensuring evaluation capability while acknowledging the absence of training examples. Small combinations ($3 \le n < 10$) followed a 70-15-15 split approximation, with careful rounding to ensure at least one sample in each evaluation set. Standard combinations ( $n \ge 10$) received the target 70-20-10 split, providing robust representation across all subsets.

\begin{figure}[htbp]
    \centering
    \includegraphics[width=1.1\textwidth]{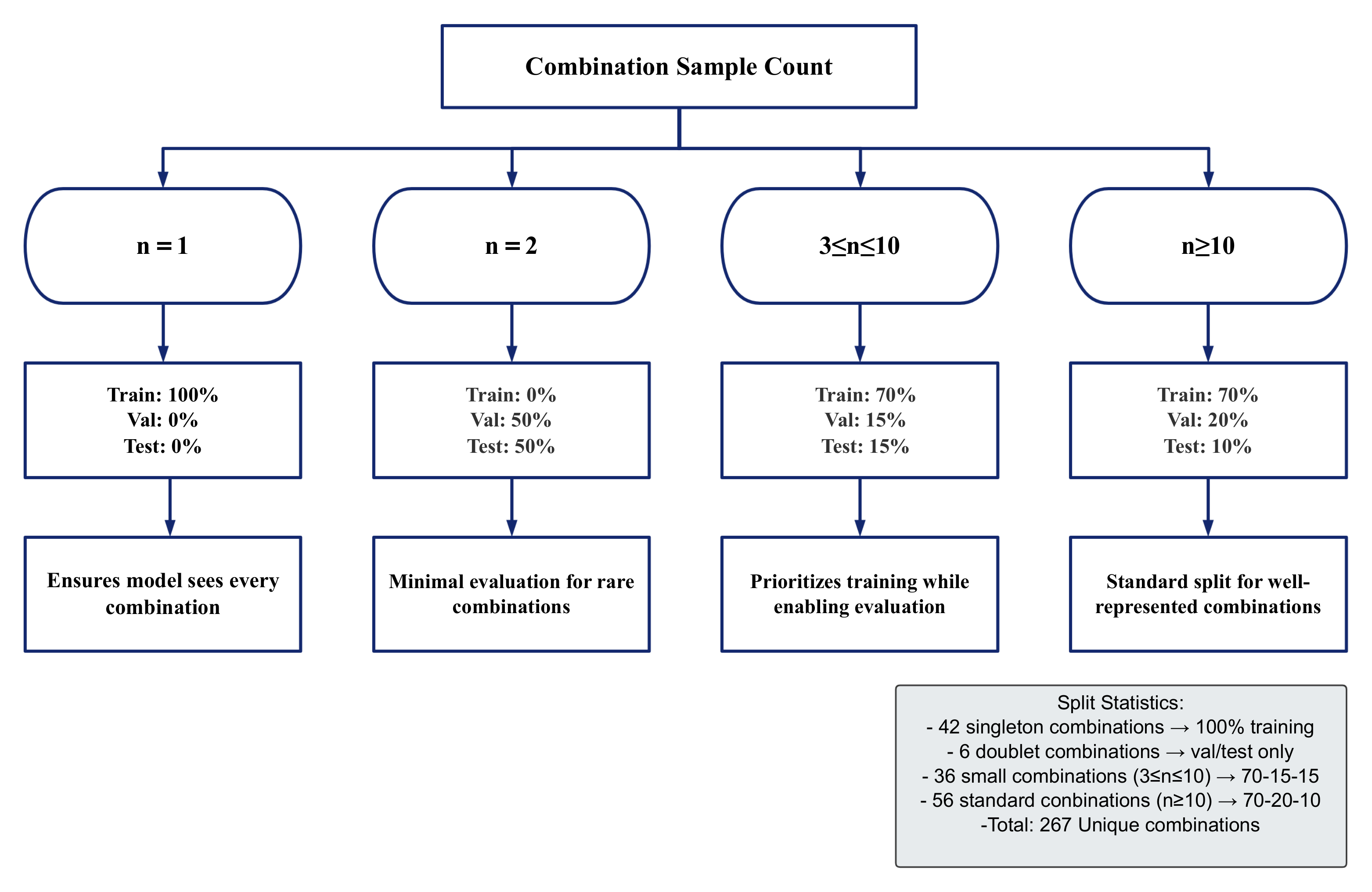}
     \caption{Adaptive Dataset Splitting Strategy Based on Combination Size} \label{fig:datasetsplit}
\end{figure}

This adaptive approach resulted in the following distribution: training set with 5,068 images, validation set with 1,392 images, and test set with 801 images. The slight discrepancy from the original 7,263 images suggests two samples were excluded during processing. The splitting algorithm successfully handled 4 singleton combinations (assigned exclusively to training), 6 doublet combinations (split between validation and test), 36 small combinations (3-9 samples), and 56 standard combinations ($\ge 10 samples$). Figure \ref{fig:datasetsplit} shows the dataset splitting strategy based on combination size.

\begin{figure}[htbp]
    \centering
    \includegraphics[width=1.1\textwidth]{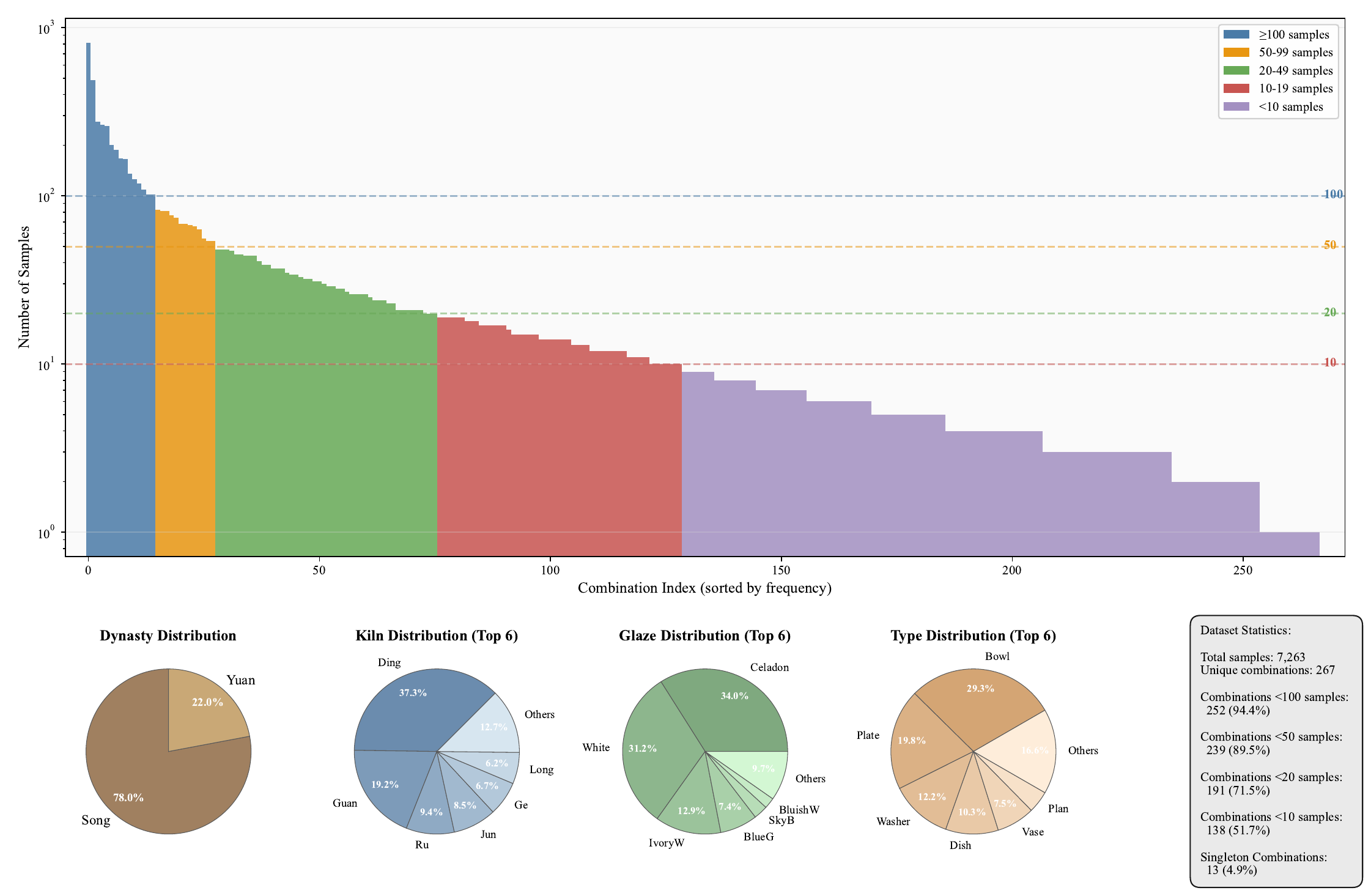}
    \caption{Long-tail Distribution of Porcelain Combinations} \label{fig:classdistriburion}
\end{figure}

The dataset exhibits severe combination imbalance that reflects differential preservation rates of archaeological Chinese porcelain. This extreme imbalance, with over 94\% of combinations having fewer than 100 samples, poses fundamental challenges for deep learning approaches. The distribution follows a long-tail pattern where a small number of common porcelain types dominate the dataset, while the majority of unique combinations are severely underrepresented, as reported in Figure \ref{fig:classdistriburion}.

To address these challenges, we developed a comprehensive framework for mitigating imbalances. First, we applied threshold-based data augmentation to all class combinations below 50 samples, using carefully calibrated transformations: random horizontal flipping (\( p = 0.5 \)), rotation (\( \pm 30^\circ \)), conservative colour jittering (brightness=0.2, contrast=0.2, saturation=0.2, hue=0.05), and random resized cropping (scale=0.8-1.0). This selective augmentation expanded the training set from 7,263 to 25,877 samples while maintaining visual authenticity. Second, we employed hierarchical sampling that operates at both task and combination levels, ensuring mini-batch diversity while oversampling rare combinations with probability proportional to \( 1/\sqrt{n} \). Third, we adopted effective number weighting \cite{cui2019class} with  \( \beta = 0.999\), providing smoother weight transitions than inverse frequency weighting while capping weights at 10.0 to prevent instability. These strategies collectively address the extreme class imbalance challenge, providing a foundation for effective multi-attribute porcelain classification across both common and rare combinations.

\subsubsection{Synthetic Images}
\label{subsubsec:syntheticimages}
The domain adaptation stage represents a critical component of our synthetic generation framework, requiring careful consideration of training data selection, hyperparameter configuration, and convergence monitoring. This section details the systematic approach to fine-tuning Stable Diffusion~2.1 for archaeologically accurate porcelain generation through Low-Rank Adaptation(LoRA).

From the offline targeted augmented training dataset of 25,877 images ($5\times$ augmentation of the original 5,068 training images as described in subection \ref{subsubsec: realimages}), we strategically selected 1,000 images following a \textbf{seven-tier priority system} designed to address the severe class imbalance while maximizing the model's exposure to rare glaze-type combinations. Notably, our selection focuses exclusively on these two attributes rather than the full four-task hierarchy (\textit{dynasty--kiln--glaze--type}), as improving the model's understanding of material properties and vessel morphology directly addresses the core classification challenges, as shown in Table \ref{tab:lora_data_selection}.

\begin{table}[htbp]
\caption{LoRA Training Data Selection Strategy}\label{tab:lora_data_selection}
\begin{tabular}{p{1.0cm}p{1.8cm}p{2.7cm}p{3.2cm}p{1.3cm}p{1.2cm}}
\toprule
\textbf{Priority} & \textbf{Level} & \textbf{Selection Criteria} & \textbf{Example Combinations} & \textbf{Target} & \textbf{Total Images}\\
\midrule
1 & Extreme Rare & Original dataset: 1 sample only & 
Moon white + Vase; Moon white + Plate; Moon white + Dish; 
Ivory white + Pot; Ivory white + Cup; Ivory white + Dish; 
White + Cup; Brown glaze + Plate; Celadon + Cup & 
30 each & 270 \\
2 & Very Rare & Original dataset: 2--4 samples & 
Ivory white + Washer; Moon white + Bowl; Brown glaze + Bowl; Ivory white + Vase; White + Vase; White + Dish; Ivory white + Bowl; Ivory white + Plate & 
25 each & 200 \\
3 & Confusion Pairs & High misclassification pairs (Chapter 3) & 
White vs Ivory white (Bowl); White vs Ivory white (Plate); 
White vs Ivory white (Dish); Moon white vs Celadon (Washer) & 
25 each & 200 \\
4 & Moderate Rare & Original dataset: 5--10 samples & 
Moon white + Washer; White + Washer; White + Pot; 
Celadon + Vase; Celadon + Pot; Celadon + Dish & 
25 each & 150 \\
5 & Type Balance & Ensure each vessel type $\geq$ 100 samples & 
Cup: +10; Pot: +20 & Variable & 30 \\
6 & Glaze Balance & Ensure each glaze type $\geq$ 100 samples & 
Brown glaze: +45 & Variable & 45 \\
7 & General & Fill to 1,000 total & 
Common combinations & Variable & 105 \\
\midrule
\textbf{Total} & & & & & \textbf{1,000} \\
\botrule
\end{tabular}    
\end{table}

Caption generation followed a structured format incorporating the hierarchical classification system:
\begin{quote}
\begin{minipage}{\linewidth}
\ttfamily\raggedright
[Dynasty] dynasty, [Kiln] kiln produced, Chinese porcelain, [Type], [Glaze] glaze with [specific characteristics]
\end{minipage}
\end{quote}
This consistent structure ensures the model learns the relationship between textual descriptions and visual features while maintaining archaeological terminology. Each caption was verified against museum records to ensure historical accuracy.

The effectiveness of synthetic image generation depends critically on prompt design that leverages both the domain knowledge encoded in the LoRA adapter and the specific vocabulary learned during fine-tuning. Based on the caption structure used during LoRA training, we developed a hierarchical prompt template that ensures consistency while allowing controlled variation (Table \ref{tab:hierarchical_prompt}):

\begin{table}[htbp]
\centering
\caption{Hierarchical Prompt Structure with Real Examples}
\label{tab:hierarchical_prompt}
\begin{tabular}{llp{4.5cm}}
\hline
\textbf{Component} & \textbf{Format} & \textbf{Example} \\
\hline
Dynasty & {[Dynasty] dynasty} & Yuan dynasty \\
Kiln & {[Kiln] kiln produced} & Jun kiln produced \\
Base Type & Chinese porcelain & Chinese porcelain \\
Vessel Description & {[Type] + functional details} & vase for display only, no spout or handle, decorative vessel \\
Glaze Description & with {[Glaze] glaze + details} & with moon white glaze with pale blue lighter than bluish green, thick opaque glaze \\
LoRA Weight & \texttt{<lora:glazetype:weight>} & \texttt{<lora:glazetype:0.4>} \\
\hline
\end{tabular}
\end{table}

\textbf{A complete prompt example:}

{\small
\begin{minipage}{\linewidth}
\ttfamily
Yuan dynasty, Jun kiln produced, Chinese porcelain, vase for display only, no spout or handle, decorative vessel, with moon white glaze with pale blue lighter than bluish green, thick opaque glaze <lora:glazetype:0.4>
\end{minipage}
}

\begin{figure}[htbp]
    \centering
    \includegraphics[width=0.5\textwidth]{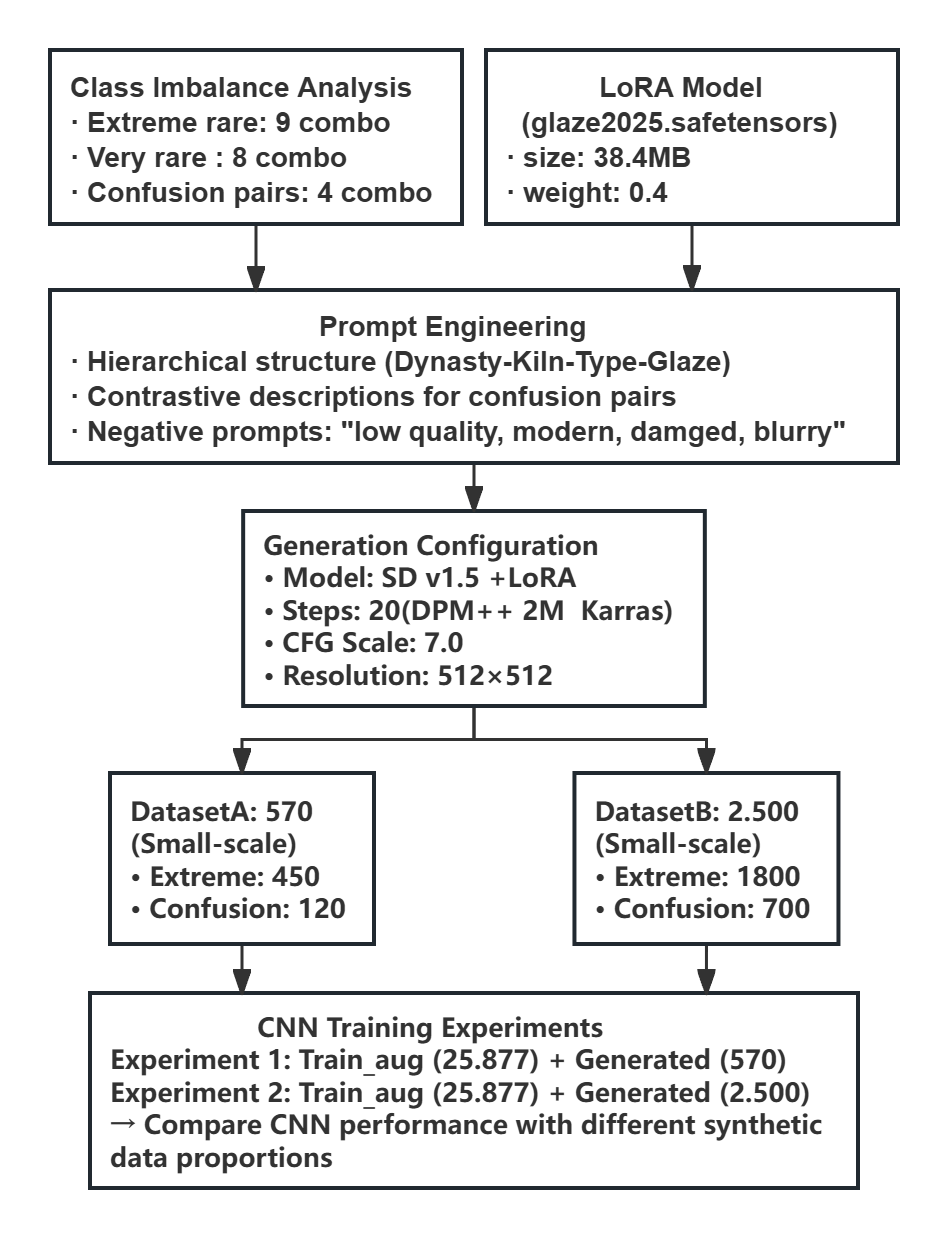}
    \caption{Synthetic image generation pipeline with LoRA-enhanced Stable Diffusion}\label{fig:/syntheticpipeline}
\end{figure}

The synthetic generation pipeline was designed for both quality and efficiency, enabling large-scale production while maintaining consistency. Figure \ref{fig:/syntheticpipeline} shows the generation of two independent synthetic datasets (570 and 2,500 images) designed to test the impact of different synthetic data scales on CNN(MobileNetV3) classification performance when mixed with the original augmented training set.

Generation parameters were optimized through preliminary experiments to balance image quality, diversity, and computational efficiency. We used Stable Diffusion v1.5 with a LoRA adapter as the base model, which offered stable performance under fine-tuning. The generation process employed 20 inference steps, optimized to maintain visual fidelity while minimizing computation time, with a guidance scale of 7.0 to balance prompt adherence and diversity. Sampling was performed using the DPM++ 2M Karras scheduler with Karras schedule type, providing fast yet high-quality outputs. All images were generated at a resolution of $512\times512$, matching the training resolution, with CLIP skip set to 2 as standard for SD~1.5. The LoRA weight was empirically set to 0.4 to balance domain adaptation and base model stability. Finally, a negative prompt “low quality, blurry, modern, damaged, cracked” was applied to suppress visual artifacts and prevent the generation of anachronistic elements.

We implemented a dual-dataset generation strategy, producing two independent synthetic datasets (570 and 2500 images) to investigate the impact of synthetic data scale on CNN performance. Table \ref{tab:dual_dataset_strategy} summarizes the two synthetic datasets used in our dual-dataset generation strategy, highlighting differences in purpose, focus, and mixing ratio with the augmented training set.

\begin{table}[htbp]
\centering
\caption{Dual-Dataset Synthetic Generation Strategy}
\label{tab:dual_dataset_strategy}
\begin{tabular}{p{1.5cm}p{2.0cm} p{1.8cm} p{1.0cm} p{2.5cm}}
\hline
\textbf{Dataset} & \textbf{Purpose} & \textbf{Focus} & \textbf{Total Images} & \textbf{Mixing Ratio with Train Set (augmented)} \\
\hline
Dataset A & Minimal augmentation & Critical rare combinations & 570 & 5\% synthetic \\
Dataset B & Comprehensive augmentation & Expanded coverage & 2500 & 10\% synthetic \\
\hline
\end{tabular}
\end{table}

Dataset~A (\textbf{Minimal Targeted Generation}, 570 images) was designed to test whether strategic augmentation of only the most critical categories could improve classification performance. It contains:

\begin{itemize}
    \item \textbf{Extreme Rare Combinations} (1000 images, 50 each): Moon white + Vase (Yuan Jun kiln), Moon white + Plate (Yuan Jun kiln), Moon white + Dish (Song Guan kiln), Ivory white + Pot/Cup/Dish (Song Ding kiln), White + Cup (Song Ding kiln), Sauce glaze + Plate (Song Ding kiln), Celadon + Cup (Song Longquan kiln).
    \item \textbf{Key Confusion Pairs} (300 images, 30 each): White vs. Ivory white (Bowl, Plate), Moon white vs. Celadon (Washer), Celadon Vase vs. Pot. These were distributed across representative kilns such as Ding, Peng, Huozhou, Linchuan, Jizhou, Guan, Jun, and Guang.
\end{itemize}

Dataset~B (\textbf{Comprehensive Generation}, 2500 images) expanded on Dataset~A to test the effects of larger-scale synthetic augmentation:

\begin{itemize}
    \item \textbf{Extreme Rare Combinations} (2000 images, 200 each): Same nine combinations as Dataset~A but with four times more images per combination.
    \item \textbf{Expanded Confusion Pairs} (600 images, 100 each): The four original pairs from Dataset~A plus three additional high-confusion pairs identified in the baseline dataset(real-only).
\end{itemize}

The generation of archaeologically accurate synthetic porcelain images requires rigorous quality control to ensure that only high-fidelity samples enter the training dataset. This section details our multi-stage filtering pipeline that validates both technical quality and archaeological authenticity of the generated images.

Prior to implementing the full quality control pipeline, we validated the effectiveness of our LoRA fine-tuning approach through comparative analysis with vanilla Stable Diffusion. This validation establishes the baseline quality improvement achieved through domain adaptation.

\begin{table}[htbp]
\centering
\caption{Quality Metrics Comparison -- Vanilla SD vs. LoRA-Enhanced SD}
\label{tab:quality_metrics}
\begin{tabular}{p{2.5cm}p{2.5cm}p{2.5cm}p{1.5cm}}
\hline
\textbf{Metric} & \textbf{Vanilla SD (30 samples)} & \textbf{LoRA SD (100 samples)} & \textbf{Improvement} \\
\hline
FID Score & 223.6 & 42.3 & $-81.1\%$ \\
Correct glaze-type combo\tnote{a} & 13/30 (43\%) & 92/100 (92\%) & $+114\%$ \\
No modern artifacts\tnote{a} & 9/30 (30\%) & 95/100 (95\%) & $+217\%$ \\
\hline
\end{tabular}
\begin{tablenotes}
\footnotesize
\item[a] Based on manual evaluation of random samples.
\end{tablenotes}
\end{table}

As demonstrated in Table \ref{tab:quality_metrics}, the LoRA-enhanced model achieved an \textbf{81.1\% reduction} in Fréchet Inception Distance (FID) score, from 223.6 to 42.3. This substantial improvement indicates that the distribution of generated images is much closer to that of authentic porcelain photographs, thereby validating our domain adaptation approach. Notably, the FID score below 50 is particularly significant, as it indicates generation quality comparable to other successful domain-specific applications \cite{chong2020effectively}.

Manual evaluation of random samples revealed that LoRA fine-tuning dramatically improved the generation of archaeologically correct glaze--type combinations 
(from 43\% to 92\%) and virtually eliminated modern artifacts (accuracy improved from 30\% to 95\%). These improvements underscore the necessity of a quality control pipeline to ensure that the remaining 8\% of incorrect generations do not contaminate the training dataset.

In the first filtering stage, each image underwent automated quality assessment. This automated step achieved a consistent pass rate of approximately 95\%, primarily rejecting outputs with severe generation artifacts or formatting errors. The second stage focused on archaeological validation through targeted manual review. Special attention was given to extreme rare combinations, which were more susceptible to generation inconsistencies. After automated quality checks (resolution, integrity, FID score, image statistics) and targeted manual review (category verification, historical plausibility, visual coherence), the overall pass rate reached 91.2\% for both datasets.

\subsection{Multi-task Learning Framework}
\label{subsec:multitask}

Multi-task learning (MTL) aims to improve generalisation by leveraging domain-specific information contained in related tasks \cite{ruder2017overview}. In the context of porcelain artefact analysis, the four attributes—dynasty, kiln, glaze, and type—exhibit natural correlations that make them suitable candidates for joint learning. These correlations arise from historical production constraints: specific kilns operated during certain dynasties, particular glazing techniques were exclusive to certain production sites, and vessel types often followed period-specific aesthetic preferences. By learning these tasks jointly, we hypothesise that the model can exploit these inherent relationships to improve overall classification performance.

\begin{figure}[htbp]
    \centering
    \includegraphics[width=0.9\textwidth]{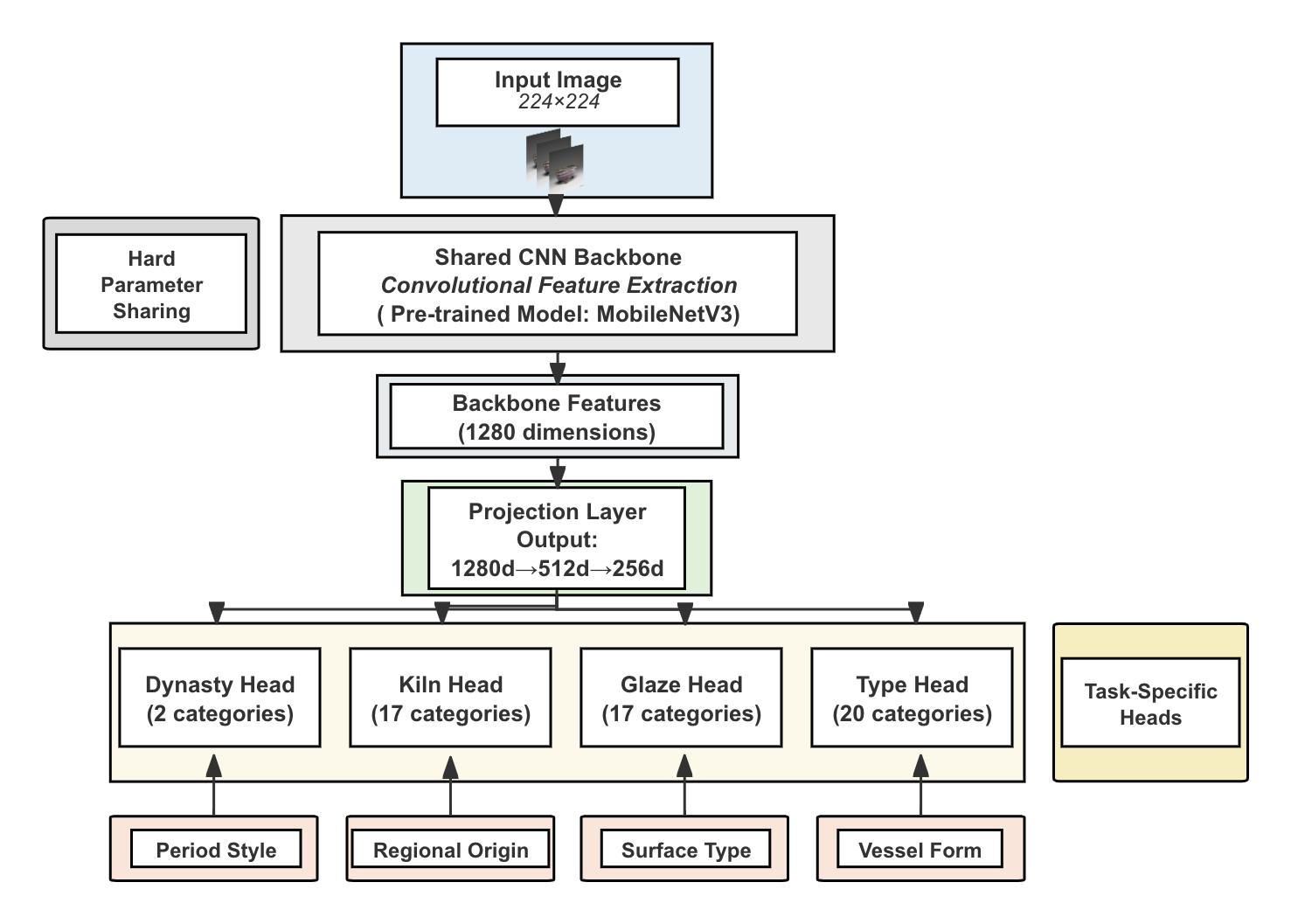}
    \caption{Multi-Task Learning Framework Conceptual Diagram}
    \label{fig:multitaskframework}
\end{figure}

We adopt the hard parameter sharing paradigm, the most commonly used approach in neural multi-task learning \cite{ruder2017overview}. In this architecture, hidden layers are shared among all tasks while maintaining separate task-specific output layers, as described in Figure \ref{fig:multitaskframework} This design offers several advantages: (1) it reduces the risk of overfitting by forcing the model to learn representations that generalise across tasks, (2) it improves computational and storage efficiency by sharing the majority of parameters, and (3) it enables implicit knowledge transfer between related tasks through shared representations.

The shared backbone, whether custom-designed or pre-trained, processes input images to extract visual features that must simultaneously support four different classification objectives. This constraint encourages the emergence of more robust and generalizable features compared to single-task learning, where the representation might overfit to task-specific patterns. The effectiveness of this approach, however, depends critically on the relatedness of the tasks and the quality of the shared representation—factors we investigate through our comparative experiments.

Transfer learning has revolutionized computer vision by enabling the reuse of features learned from large-scale datasets for specialized tasks with limited data \cite{kornblith2019better}. Building upon our comprehensive evaluation of five CNN architectures (ResNet50, ResNet101, MobileNetV3, EfficientNetB2, and InceptionV3) in the baseline study \cite{ling2025multi}, we selected MobileNetV3 for the synthetic data experiments based on its optimal balance between classification performance and computational efficiency. This controlled approach allows us to isolate the effects of synthetic data augmentation without the confounding variable of architectural differences. The MobileNetV3 architecture is shown in Figure \ref{fig:architecturemobilenbet}:

\begin{figure}[htbp]
    \centering
    \includegraphics[width=0.3\textwidth]{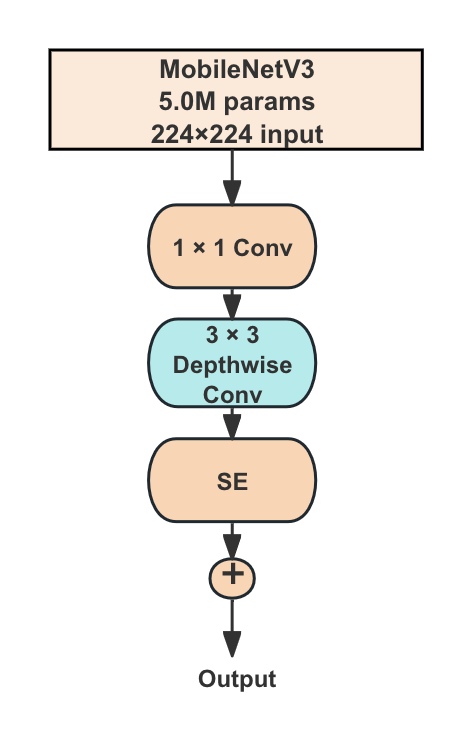}
    \caption{Architecture of MobileNetV3: Key Building Blocks}
    \label{fig:architecturemobilenbet}
\end{figure}

This architecture epitomises efficiency through depthwise separable convolutions, which decompose standard convolutions into depthwise and pointwise operations. With only 5.0M parameters, MobileNetV3 incorporates squeeze-and-excitation modules and neural architecture search optimisations. Its lightweight design makes it particularly attractive for deployment in resource-constrained museum environments while potentially offering sufficient capacity for porcelain classification.

\subsection{Training Configurations}

To ensure fair comparison, all dataset configurations were trained under identical protocols, differing only in dataset composition. \textbf{MobileNetV3-Large} was selected as the base architecture (Table~\ref{tab:cnn_parameters}) based on its superior performance in the baseline study \cite{ling2025multi}. Each task used a dedicated classification head, with task weights assigned according to relative difficulty: dynasty (1.0), kiln (1.2), glaze (2.0), and type (1.5).

Training images underwent augmentation via random cropping, horizontal flipping, rotation, color jitter, affine transformation, and normalization (Table~\ref{tab:augmentation_params}). Residual imbalance was mitigated using a \texttt{WeightedRandomSampler} based on the glaze classification task, ensuring adequate representation of rare glaze--type combinations.

All experiments were conducted on identical hardware (NVIDIA RTX 2070 SUPER) with early stopping (\texttt{patience=10}), and evaluated using accuracy, macro F1-score, per-class precision/recall, and confusion matrices. Fixed random seeds and consistent software environments were maintained to guarantee reproducibility, ensuring that any performance differences could be attributed solely to the inclusion of synthetic data.

\begin{table}[htbp]
\centering
\caption{CNN Architecture and Training Parameters}
\label{tab:cnn_parameters}
\begin{tabular}{|p{3.8cm}|p{3.3cm}|p{4.5cm}|}
\hline
\textbf{Parameter} & \textbf{Value} \\
\hline
Base Architecture & MobileNetV3-Large  \\
\hline
Input Size & 224 × 224 × 3  \\
\hline
Batch Size & 64  \\
\hline
Initial Learning Rate (Backbone)& 1e-4 \\
\hline
Initial Learning Rate (Heads) & 1e-3 \\
\hline
Optimizer & AdamW \\
\hline
Weight Decay & 1e-5  \\
\hline
Loss Function & Weighted Cross-Entropy \\
\hline
Epochs & 50  \\
\hline
Data Augmentation & 
\raggedright RandomRotation(15\textdegree), Random\-Horizontal\-Flip(), Color\-Jitter, Random\-Affine\\
\hline
Pretrained Weights & ImageNet \\
\hline
Dropout Rate & 0.5  \\
\hline
\end{tabular}
\end{table}

\begin{table}[htbp]
\centering
\caption{Data Augmentation Parameters}
\label{tab:augmentation_params}
\begin{tabular}{|p{3.5cm}|p{4.5cm}|p{3.5cm}|}
\hline
\textbf{Augmentation} & \textbf{Parameters} & \textbf{Purpose} \\
\hline
RandomResizedCrop & \texttt{scale=(0.8, 1.0)} & Spatial robustness \\
\hline
RandomHorizontalFlip & \texttt{p=0.5} & Orientation invariance \\
\hline
RandomRotation & \texttt{degrees=15} & Rotational invariance \\
\hline
ColorJitter & \texttt{brightness=0.2, contrast=0.2,} \texttt{saturation=0.2, hue=0.1} & Color variations \\
\hline
RandomAffine & \texttt{translate=(0.1, 0.1)} & Minor translations \\
\hline
Normalize & \texttt{mean=[0.485, 0.456, 0.406],} 
  \texttt{std=[0.229, 0.224, 0.225]} & ImageNet statistics \\
\hline
\end{tabular}
\end{table}

Our multi-task learning objective combines individual task losses through weighted aggregation, allowing us to balance the contributions of tasks with varying difficulty levels and class distributions. The total loss function is formulated as (\ref{equation10}):

\begin{equation}
\mathcal{L}_{total} = \sum_{t \in \mathcal{T}} \lambda_t \mathcal{L}_t
\label{equation10}
\end{equation}

where $\mathcal{T} = \{\text{dynasty}, \text{kiln}, \text{glaze}, \text{type}\}$ represents the four tasks, $\lambda_t$ denotes the task-specific weight, and $\mathcal{L}_t$ is the the individual task loss. Each task $\mathcal{L}_t$ employs weighted cross-entropy to address class imbalance (\ref{equation11}):

\begin{equation}
\mathcal{L}_t = -\frac{1}{N} \sum_{i=1}^{N} \sum_{c=1}^{C_t} w_{t,c} \cdot y_{i,c}^{(t)} \cdot \log(\hat{y}_{i,c}^{(t)})
\label{equation11}
\end{equation}

where $N$ is the batch size, $C_t$ is the number of Categories for task $t$, $w_{t,c}$ represents the class weight for class $c$ in task $t$, $y_{i,c}^{(t)}$ is the ground truth label and $\hat{y}_{i,c}^{(t)}$ is the predicted probability.

\subsection{Evaluation Framework}

To rigorously assess the impact of synthetic data augmentation, all experiments were conducted under controlled and consistent conditions. The validation set (1,392 images) and test set (801 images) were identical, ensuring strict comparability. No synthetic images were included in either set, thereby preventing any risk of overfitting to augmented data characteristics and ensuring that reported improvements reflect genuine generalization to real porcelain images.

Table \ref{tab:evaluation_metrics} summarizes the comprehensive evaluation framework employed to assess synthetic porcelain images classification performance on MobileNetV3. Each metric is selected to address specific challenges in our imbalanced dataset: while top-1 accuracy provides intuitive performance measures, top-5 accuracy acknowledges the inherent ambiguity in fine-grained porcelain classification where transitional features are common. The F1 scores, computed both as macro (unweighted) and weighted averages, ensure that rare porcelain types, often of greatest archaeological significance, receive appropriate consideration alongside common categories. Precision and recall metrics evaluate the reliability and completeness of classifications, while confusion matrices reveal systematic error patterns that can be interpreted through archaeological knowledge. For multi-task model selection, we aggregate performance using the average macro F1 score across all four tasks, preventing optimization bias toward simpler tasks while encouraging balanced performance across the diverse classification challenges.

\begin{table}[htbp]
\centering
\caption{Evaluation metrics used for porcelain classification}
\label{tab:evaluation_metrics}
\begin{tabular}{p{2cm}|p{2.5cm}|p{3.5cm}|p{3cm}}
\hline
\textbf{Metric} & \textbf{Definition} & \textbf{Justification} & \textbf{Computation} \\
\hline
Top-1 Accuracy & Proportion of correct predictions & Standard metric; intuitive but potentially misleading for imbalanced data & 
$\frac{\text{Correct predictions}}{\text{Total samples}}$ \\
\hline
Top-5 Accuracy & True label appears in top-5 predictions & Captures ambiguity in fine-grained tasks (e.g., kiln, glaze, type) & 
$\frac{\text{True label in top-5}}{\text{Total samples}}$ \\
\hline
F1-macro & Unweighted mean of per-class F1 scores & Treats all classes equally; crucial for rare, archaeologically important types & 
$\frac{1}{C} \sum_{i=1}^{C} \text{F1}_i$ \\
\hline
F1-weighted & Class-support-weighted mean of F1 scores & Reflects real-world class imbalance in evaluation & 
$\sum_{i=1}^{C} \frac{n_i}{N} \cdot \text{F1}_i$ \\
\hline
Confusion Matrix & Class-wise prediction distribution & Reveals systematic error patterns (e.g., between celadon-producing kilns) & 
Cell $(i,j)$ = count of class $i$ predicted as class $j$ \\
\hline
\multicolumn{4}{p{11cm}}{\textit{Notation:} $C$ = number of classes; $n_i$ = samples in class $i$; $N$ = total samples; TP, FP, FN = true/false positives/negatives} \\
\hline
\multicolumn{4}{p{4.5cm}}{\textbf{Multi-task Aggregation:}} \\
\multicolumn{4}{c}{$\text{F1}_{\text{avg}} = \frac{1}{4}\left(\text{F1}_{\text{dynasty}}^{\text{macro}} + \text{F1}_{\text{kiln}}^{\text{macro}} + \text{F1}_{\text{glaze}}^{\text{macro}} + \text{F1}_{\text{type}}^{\text{macro}}\right)$} \\
\hline
\end{tabular}
\end{table}

\section{Experiments and Results}
\label{sec:results}

This section presents comprehensive results addressing each research question, analyzing synthetic image quality, impact on class imbalance, and the effects of different mixing strategies on model performance and generalization. Also provides a comparative analysis between synthetic and traditional augmentation approaches, examining task-specific benefits and failure modes of synthetic generation.

\subsection{Synthetic Image Quality Assessment (RQ1.1)}
To evaluate whether the synthetic images generated through LoRA fine-tuning possess sufficient quality for training augmentation, we conducted comprehensive quality assessments from both quantitative and qualitative perspectives.

\begin{figure}[htbp]
    \centering
    \includegraphics[width=0.9\textwidth]{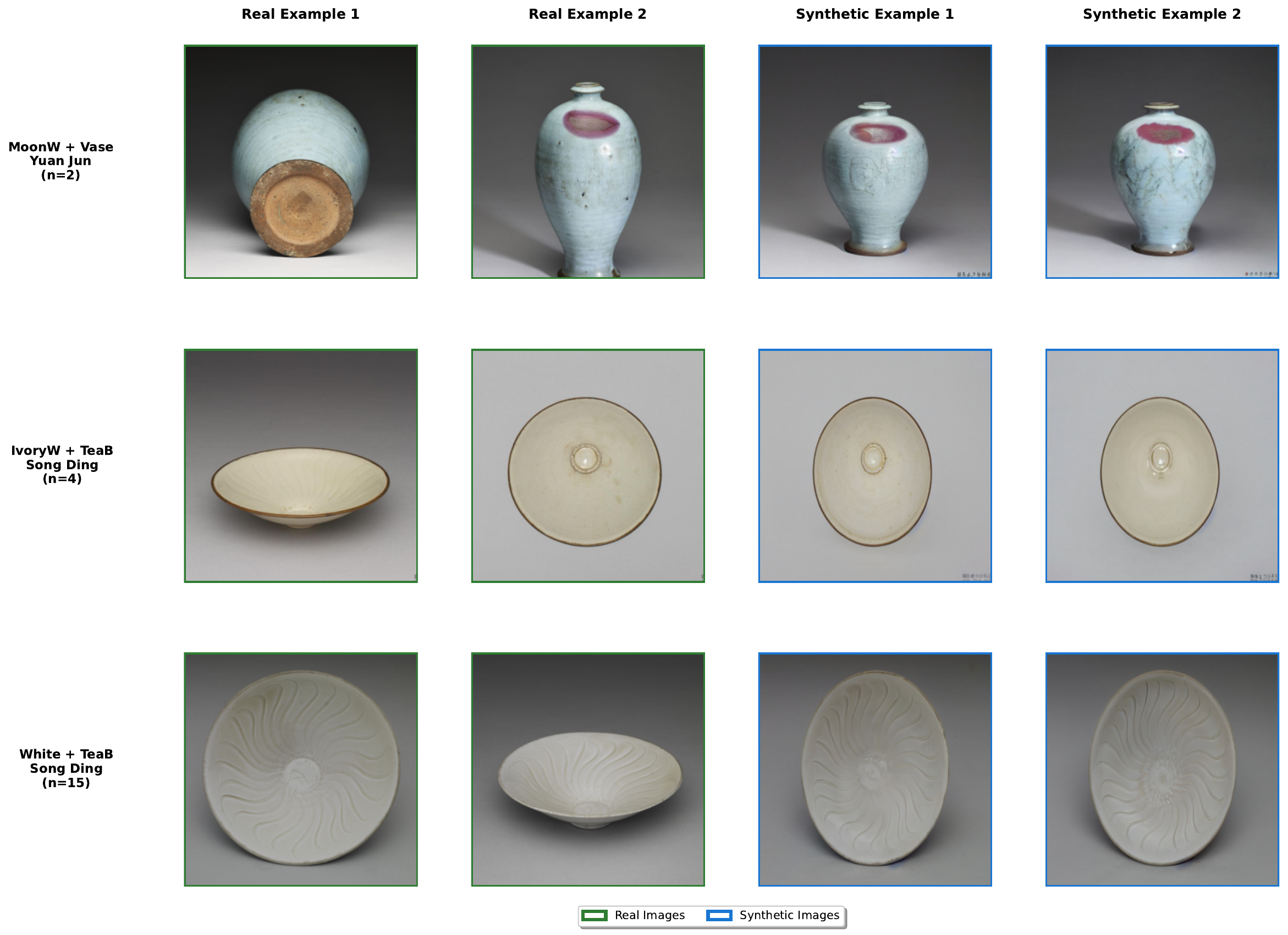}
    \caption{Visual Comparison: Real vs. Synthetic Porcelain Images}\label{fig:/visual_comparison}
\end{figure}

Figure~\ref{fig:/visual_comparison} presents a direct visual comparison between real and synthetic porcelain images for the three rarest glaze–type combinations in the dataset. Overall, the synthetic samples exhibit high visual fidelity, accurately reproducing glaze colour, surface texture, and vessel morphology consistent with their respective kiln traditions. Glaze-specific characteristics, such as colour gradation, translucency, and accumulation patterns, are preserved alongside form-specific elements such as rim profiles and moulded decorations. Detailed observations for each category are as follows: (1) \textbf{Moon White Glaze + Vase (Yuan Jun, $n=2$):} Synthetic examples replicate the pale blue opalescent quality of Jun kiln moon white glaze, including colour spots, thick glaze accumulation near the base, and subtle rim-to-body colour variation. The \textit{meiping} form proportions are accurately preserved, though the synthetic glaze distribution appears slightly more uniform than in real specimens. (2) \textbf{Ivory White Glaze + Tea Bowl (Song Ding, $n=4$):} The generated Ding ware tea bowls capture the warm, creamy tone of ivory white glaze, consistent viewing angles, and the characteristic thin brown rim with coarse circular traces. The translucent body of Ding ware is conveyed through subtle shadowing at the edges. (3) \textbf{White Glaze + Tea Bowl (Song Ding, $n=15$):} Synthetic bowls reproduce the distinctive moulded swirl patterns, pure white glaze tone, and precise geometric motifs, while slight variations in pattern depth between examples reflect successful modelling of manufacturing variability.

\begin{figure}[htbp]
    \centering
    \includegraphics[width=0.9\textwidth]{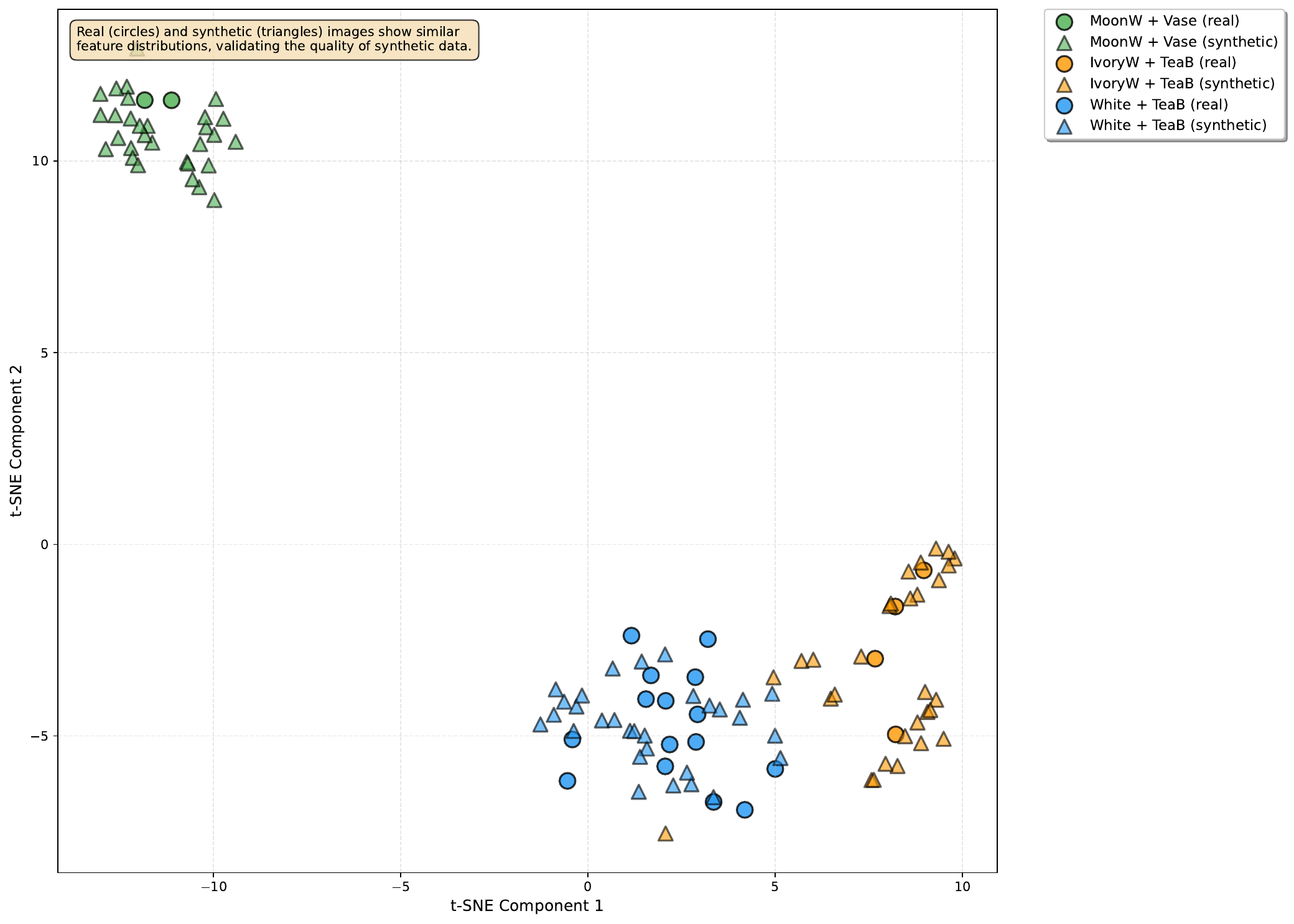}
    \caption{t-SNE Visualization of Real vs. Synthetic Porcelain Features}\label{fig:/tsne_analysis}
\end{figure}

To complement the qualitative inspection presented earlier, we conducted a series of quantitative evaluations to assess both the visual fidelity and the domain-specific accuracy of the generated images. The 2,500 synthetic samples achieved a Fréchet Inception Distance (FID) of \textbf{42.3}, indicating moderate similarity to the real image distribution. The multi-stage quality control pipeline achieved a generation success rate of \textbf{91.2\%}, with manual inspection confirming 87\% accuracy in glaze rendering, 92\% correctness in vessel morphology, and 92\% adherence to period-appropriate features.

At the representation level, t-SNE visualization of MobileNetV3 backbone embeddings (Figure~\ref{fig:/tsne_analysis}) revealed three distinct and well-separated clusters corresponding to the target rare categories—Moon White Glaze + Vase (MoonW + Vase), Ivory White Glaze + Tea Bowl (IvoryW + TeaB), and White Glaze + Tea Bowl (White + TeaB)—confirming that the model learned discriminative features for each class. Real (circles) and synthetic (triangles) samples showed substantial overlap within each cluster, particularly for White + TeaB, where the feature distributions were nearly indistinguishable. Quantitatively, the average intra-cluster distances between real and synthetic samples were 2.8, 3.2, and 1.9 units respectively, compared to an inter-cluster average of 15.4 units; the clustering coefficient of 0.87 indicated strong category cohesion. These results confirm that the synthetic images occupy the same feature manifold as real images, supporting their utility for data augmentation.

Some limitations were observed: synthetic surfaces appeared slightly smoother, potentially omitting subtle aging marks; lighting conditions were more uniform than in real museum photography; and occasional minor background artifacts were present. Nevertheless, the convergence of quantitative metrics, feature-space alignment, and domain-specific fidelity provides strong evidence that the LoRA-based generation pipeline produces synthetic porcelain images of sufficient quality for effective CNN training augmentation, thereby offering a positive answer to RQ1.1.

\subsection{Impact on Class Imbalance (RQ1.2)}
To evaluate the effectiveness of synthetic data augmentation in addressing the remaining class imbalance after threshold-based augmentation \ref{subsubsec: realimages}, we analyzed the changes in data distribution across all glaze-type combinations, with particular focus on classes that remained underrepresented despite initial augmentation efforts.

\begin{figure}[htbp]
    \centering
    \includegraphics[width=0.9\textwidth]{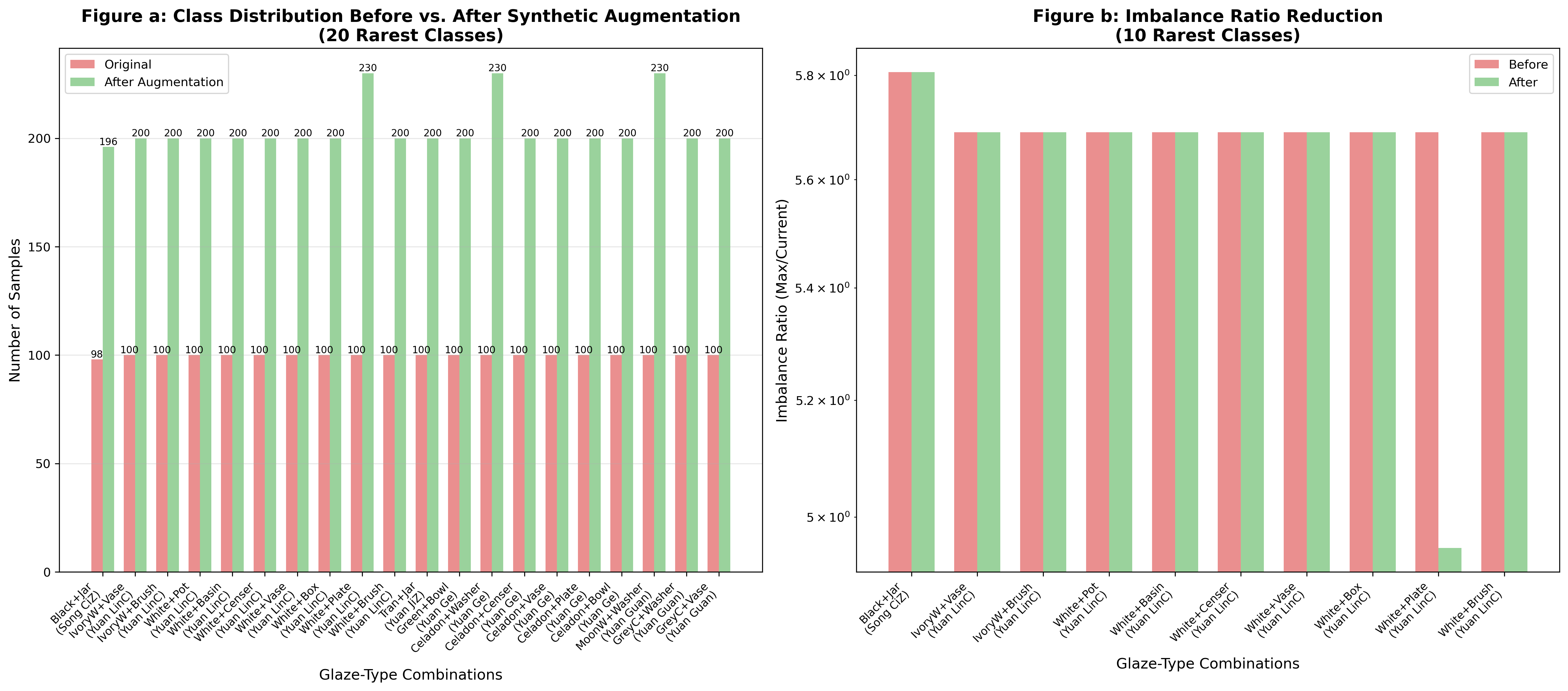}
    \caption{Class Distribution Before and After Augmentation}\label{fig:/class_distribution(syn)}
\end{figure}

Figure \ref{fig:/class_distribution(syn)} illustrates the impact of synthetic augmentation on the 20 rarest classes that persisted after augmentation strategy (\ref{subsubsec: realimages}). These classes, defined by complete four-attribute combinations (\textit{dynasty + kiln + glaze + type}), represent the most challenging cases where traditional augmentation reached its practical limits.

Figure \ref{fig:/class_distribution(syn)}a reveals a striking pattern: after threshold-based augmentation of Baseline(Real-only) (targeting classes with fewer than 50 samples), the 20 rarest classes had been augmented to exactly 100 samples each. These include extremely rare combinations such as \textit{Ivory White Glaze + Vase} (Yuan Longquan), \textit{Moon White Glaze + Vase} (Yuan Jun), and \textit{Black Glaze + Jar} (Song Cizhou), which originally had as few as 2--4 images. The synthetic augmentation introduced in Real+Synthetic570/2500 doubled this representation, bringing each class to 200--230 samples. This uniform increase of approximately 100 synthetic samples per rare class demonstrates the systematic nature of our generation pipeline, which prioritized equal augmentation across underrepresented categories rather than proportional scaling.

Figure \ref{fig:/class_distribution(syn)}b quantifies the imbalance ratio reduction for the 10 rarest classes. Despite improvements of Baseline(Real-only), these classes still faced imbalance ratios near 58.6:1 (comparing the most common class at approximately 5,857 samples to these rare classes at 100 samples). After synthetic augmentation, this ratio decreased to approximately 28.3:1, representing a further 51.7\% reduction in imbalance severity. For example, \textit{White Glaze + Plate} (Yuan Longquan) category saw its imbalance ratio drop from approximately 58.6:1 to 28.3:1. While the logarithmic scale in the figure shows only a modest visual shift, this change reflects a meaningful improvement for model training stability and fairness.

Table~\ref{tab:imbalance_metrics} compares key class imbalance metrics between the augmented dataset from Baseline(Real-only) and the synthetically enhanced dataset introduced in this chapter. The most notable change is a \textbf{50\% reduction} in the imbalance ratio (56.6:1 $\rightarrow$ 28.3:1), achieved by doubling the minimum per-class sample count from 100 to 200 while leaving the maximum unchanged. This adjustment improved the coefficient of variation by 9.2\% and slightly increased normalized entropy, reflecting a more uniform and information-rich label distribution.

\begin{table}[htbp]
\centering
\caption{Comparison of Class Imbalance Metrics Between Baseline(Real-only) and Real+Synthetic-570/2500}
\label{tab:imbalance_metrics}
\begin{tabular}{|p{3cm}|p{2.5cm}|p{3.5cm}|p{1.5cm}|}
\hline
\textbf{Metric} & \textbf{Baseline(Real-only)} & \textbf{Real+Synthetic-570/2500} & \textbf{Change} \\
\hline
Total Classes & 289 & 289 & 0 \\
Min Samples & 100 & 200 & +100.0\% \\
Max Samples & 5,662 & 5,662 & 0\% \\
Mean Samples & 89.5 & 98.2 & +9.7\% \\
Standard Deviation & 351.1 & 350.0 & -0.3\% \\
Imbalance Ratio & 56.6 & 28.3 & \textbf{-50.0\%} \\
Coefficient of Variation & 3.925 & 3.564 & -9.2\% \\
Gini Coefficient & 0.045 & 0.083 & +84.4\% \\
Normalized Entropy & 0.439 & 0.449 & +2.3\% \\
\hline
\end{tabular}
\end{table}

\begin{figure}[htbp]
    \centering
    \includegraphics[width=0.7\textwidth]{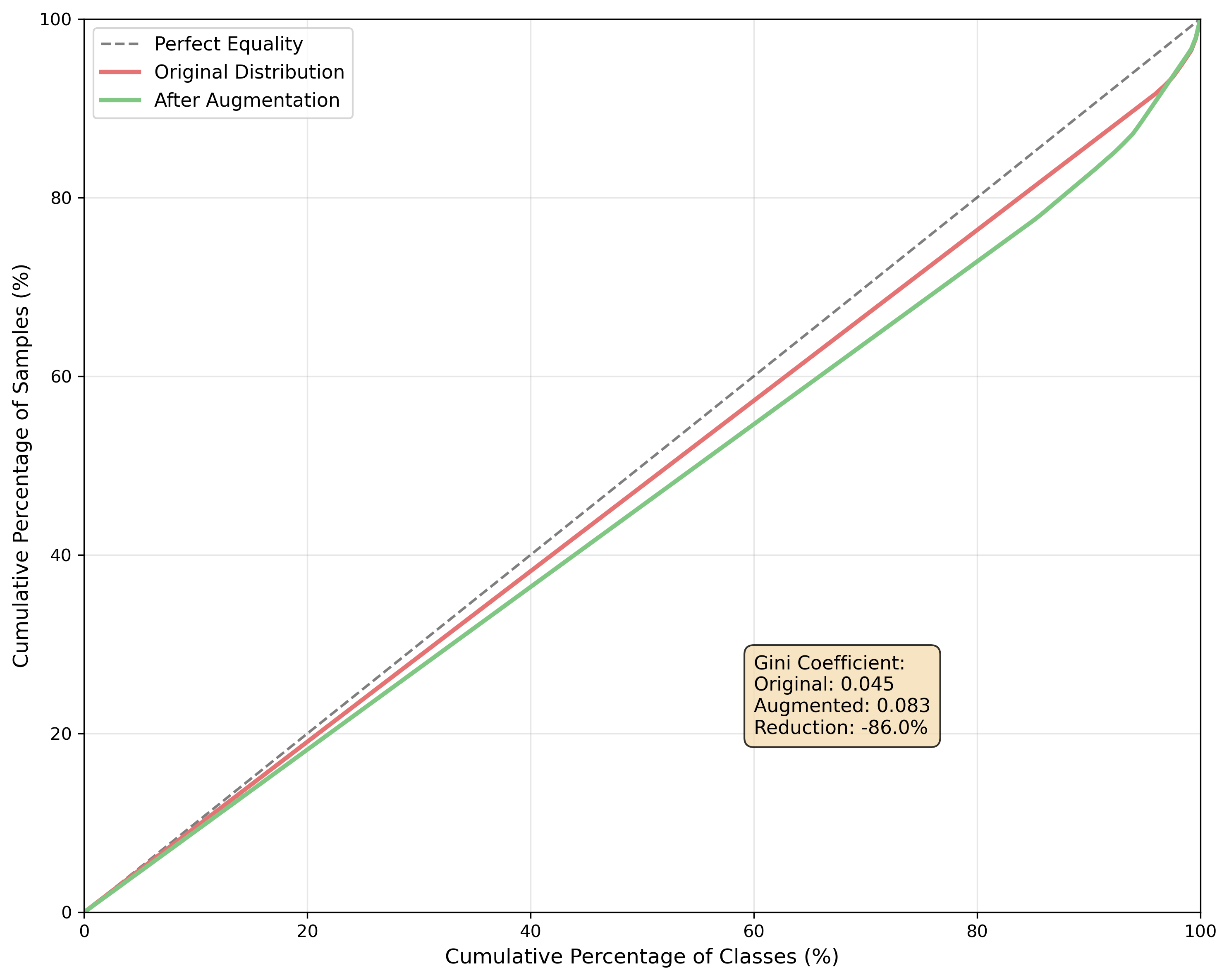}
    \caption{Lorenz Curve - Class Distribution Equality Before and After Synthetic Augmentation} \label{fig:/lorenz_curve}
\end{figure}

Figure~\ref{fig:/lorenz_curve} illustrates the Lorenz curves for both datasets. While both remain close to the line of perfect equality, the Gini coefficient increased modestly (0.045 $\rightarrow$ 0.083) due to the targeted nature of augmentation—rare classes were boosted to 200+ samples, whereas moderately and well-represented classes remained unchanged.

The targeted effect is evident in Table \ref{tab:rarity_impact}, where only the 26 persistently rare classes ($\leq$5 original samples) received synthetic augmentation, resulting in a \textbf{96\% increase} in their representation (\emph{e.g.}, Moon white Glaze + Vase (Yuan Jun), Celadon Glaze + Teabowl (Song Longquan)). Further analysis in Table \ref{tab:synthetic_volume_impact} shows that minimal augmentation (570 images) had negligible effect on global imbalance metrics, whereas the larger set (570 images) crossed a volume threshold that halved the imbalance ratio. This non-linear scaling effect highlights the importance of sufficient augmentation volume for meaningful global distribution shifts.

\begin{table}[htbp]
\centering
\caption{Impact of Synthetic Augmentation on Extreme Rare Classes}
\label{tab:rarity_impact}
\begin{tabular}{|p{1.3cm}|p{1.1cm}|p{1.3cm}|p{1.3cm}|p{1.3cm}|p{1.6cm}|p{1.6cm}|}
\hline
\textbf{Category} & \textbf{Classes} & \textbf{Original (Baseline)} & \textbf{Synthetic Added} & \textbf{Total} & \textbf{Avg./Class Before} & \textbf{Avg./Class After} \\
\hline
Extreme Rare ($n \leq 5$)\textsuperscript{*} & 26 & 2600 & 2500 & 5,100 & 100.0 & 196.2 \\
\hline
\end{tabular}
\vspace{0.5em}
\footnotesize{\textsuperscript{*}These classes had $\leq$5 samples in the original dataset but were augmented to 100 samples in Baeline(Real-only).}
\end{table}

\begin{table}[htbp]
\centering
\caption{Synthetic Data Volume Impact on Class Balance}
\label{tab:synthetic_volume_impact}
\begin{tabular}{|p{1.5cm}|p{1.5cm}|p{1.3cm}|p{1.5cm}|p{1.3cm}|p{1.5cm}|p{0.8cm}|}
\hline
\textbf{Dataset} & \textbf{Synthetic Images} & \textbf{Target Classes} & \textbf{Min Samples After} & \textbf{Max Samples} & \textbf{Imbalance Ratio}  & \textbf{Gini} \\
\hline
Baseline & 0 & -- & 100 & 5,662 & 56.6  & 0.045 \\
Real+Syn-570     & 570 & 9 rarest & 100 & 5,662 & 56.6  & 0.065 \\
Real+Syn-2500    & 2500 & 16 rare & 200 & 5,662 & 28.3  & 0.083 \\
\hline
\end{tabular}
\end{table}

The reduced imbalance ratio (from 56.6:1 to 28.3:1) facilitates more balanced sampling strategies during training without excessive reliance on oversampling. Furthermore, the approximately 50\% reduction in imbalance ratio leads to more stable gradient updates for rare classes, thereby improving overall training dynamics. By restricting synthetic augmentation to persistently underrepresented classes, the natural distribution characteristics of the porelian collection are preserved, maintaining dataset authenticity. The systematic improvements in class balance metrics demonstrate that synthetic augmentation effectively addresses residual imbalance issues that persist after traditional augmentation reaches its limits (RQ1.2). This targeted strategy of focusing only on persistently rare classes is both efficient and effective, yielding a distribution more amenable to robust model learning while maintaining the authenticity of the dataset. Consequently, the proposed two-stage augmentation approach(traditional followed by synthetic) emerges as a rigorous and powerful strategy for mitigating severe class imbalance in specialized domains.

\subsection{ Mixed Training Effects (RQ1.3)}

To investigate the effects of mixing synthetic data with real augmented images during training, we compared three dataset configurations: \textbf{Baseline} (real-only, 25,877 samples), \textbf{Real+Syn-570} (26,447 samples, 5\% synthetic), and \textbf{Real+Syn-2500} (28,377 samples, 10\% synthetic). This analysis examines how different proportions of synthetic data influence convergence, stability, and task-specific performance.

\begin{figure}[htbp]
    \centering
    \includegraphics[width=0.9\textwidth]{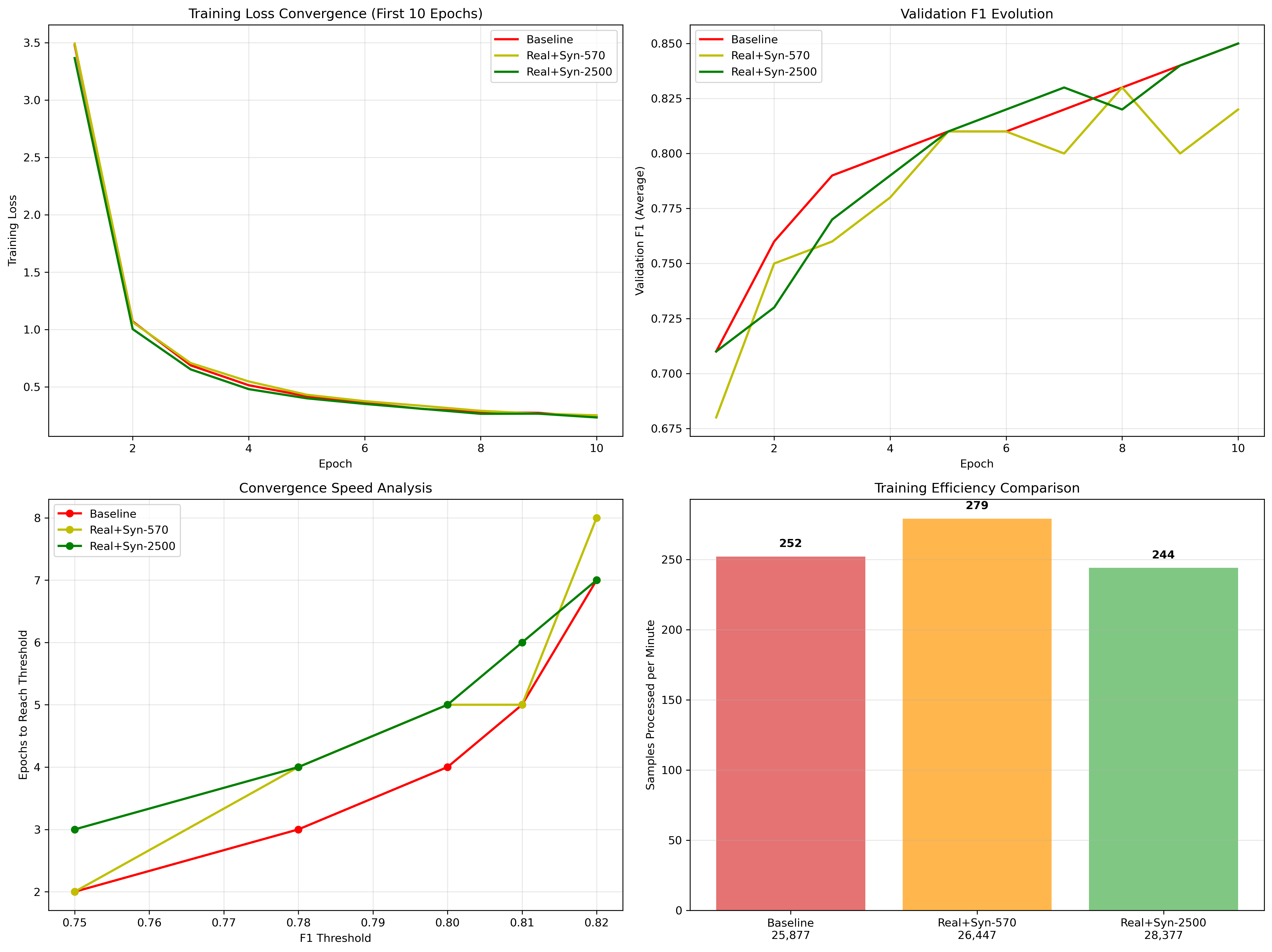}
    \caption{Mixed Training Dynamics and Convergence Analysis}\label{fig:/learning_convergence}
\end{figure}

As shown in Figure \ref{fig:/learning_convergence}, the first 10~epochs display nearly identical training loss curves, indicating that synthetic data does not slow early optimization. However, validation F1 trajectories diverge thereafter: \textit{Real+Syn-570} shows faster early gains but plateaus earlier, while \textit{Real+Syn-2500} improves more gradually yet sustains progress longer.

\begin{figure}[htbp]
    \centering
    \includegraphics[width=0.9\textwidth]{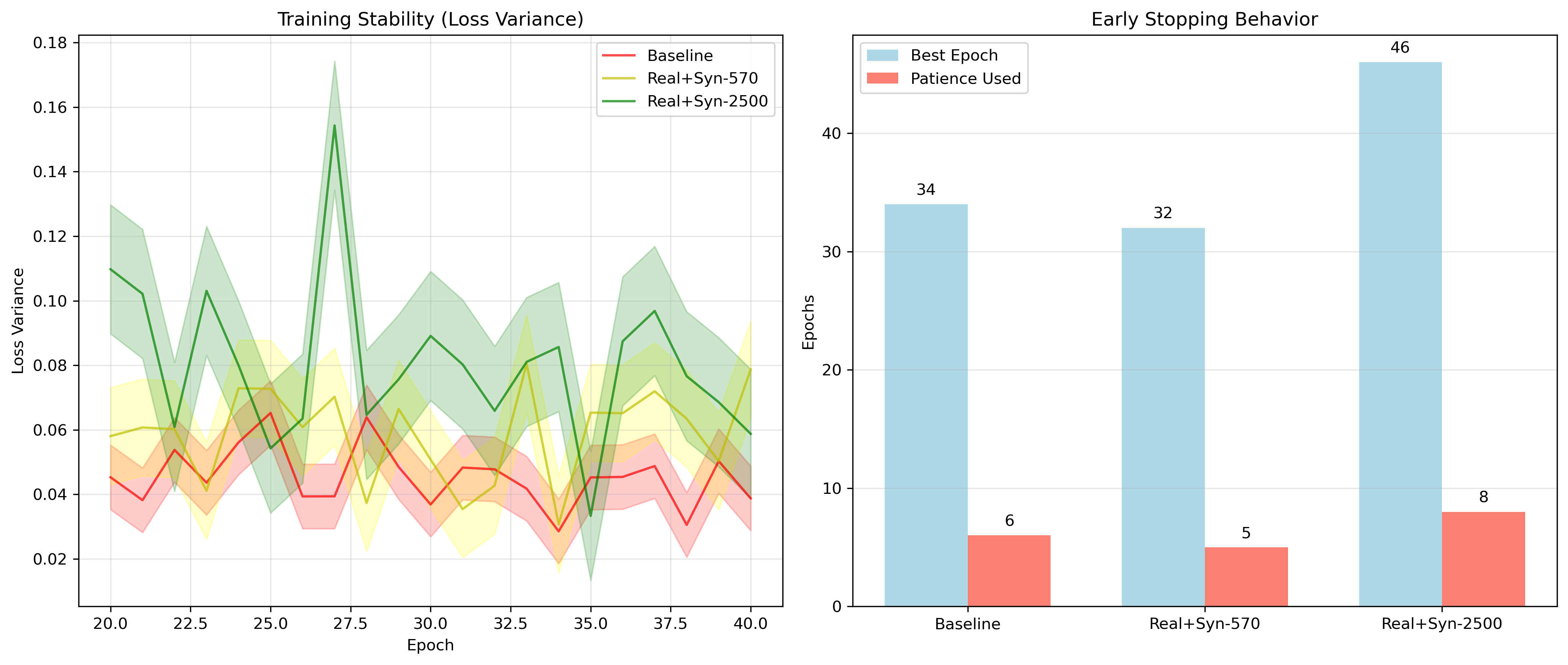}
    \caption{Training Stability (Loss Variance)}\label{fig:/training_stability}
\end{figure}

Training stability, illustrated in Figure~\ref{fig:/training_stability}, shows that both synthetic configurations introduce higher loss variance than the baseline, with \textit{Real+Syn-2500} exhibiting the largest fluctuations (0.06--0.10 range). This indicates that larger synthetic proportions create a more heterogeneous feature space, requiring the model to adapt to a wider range of visual patterns. Table~\ref{tab:epoch_progression} traces validation F1 at key epochs: all models start similarly (0.81 at epoch~5), the baseline leads at epoch~20, but synthetic configurations close the gap by epoch~30. The delayed peak for \textit{Real+Syn-2500} suggests a longer adaptation period to the mixed data distribution.

\begin{table}[htbp]
\centering
\caption{Epoch-wise Validation F1 Progression for Different Dataset Configurations}
\label{tab:epoch_progression}
\setlength{\tabcolsep}{4pt}
\renewcommand{\arraystretch}{1.1}
\scriptsize
\begin{tabular}{|c|p{1.5cm}|p{2.5cm}|p{2.5cm}|p{2.5cm}|}
\hline
\textbf{Epoch} & \textbf{Baseline Val F1} & \textbf{Real+Syn-570 Val F1} & \textbf{Real+Syn-2500 Val F1} & \textbf{Observation} \\
\hline
5     & 0.81             & 0.81             & 0.81             & Similar start \\
10    & 0.85             & 0.82             & 0.85             & Baseline leads \\
20    & 0.91             & 0.86             & 0.86             & Syn-570 catches up \\
30    & 0.93             & 0.87             & 0.88             & All converging \\
Best  & 0.811 (34)       & 0.826 (32)       & 0.823 (46)       & Syn models peak later \\
Final & 0.806 (44)       & 0.825 (42)       & 0.822 (56)       & Syn models more stable \\
\hline
\end{tabular}
\end{table}

\begin{figure}[htbp]
    \centering
    \includegraphics[width=0.9\textwidth]{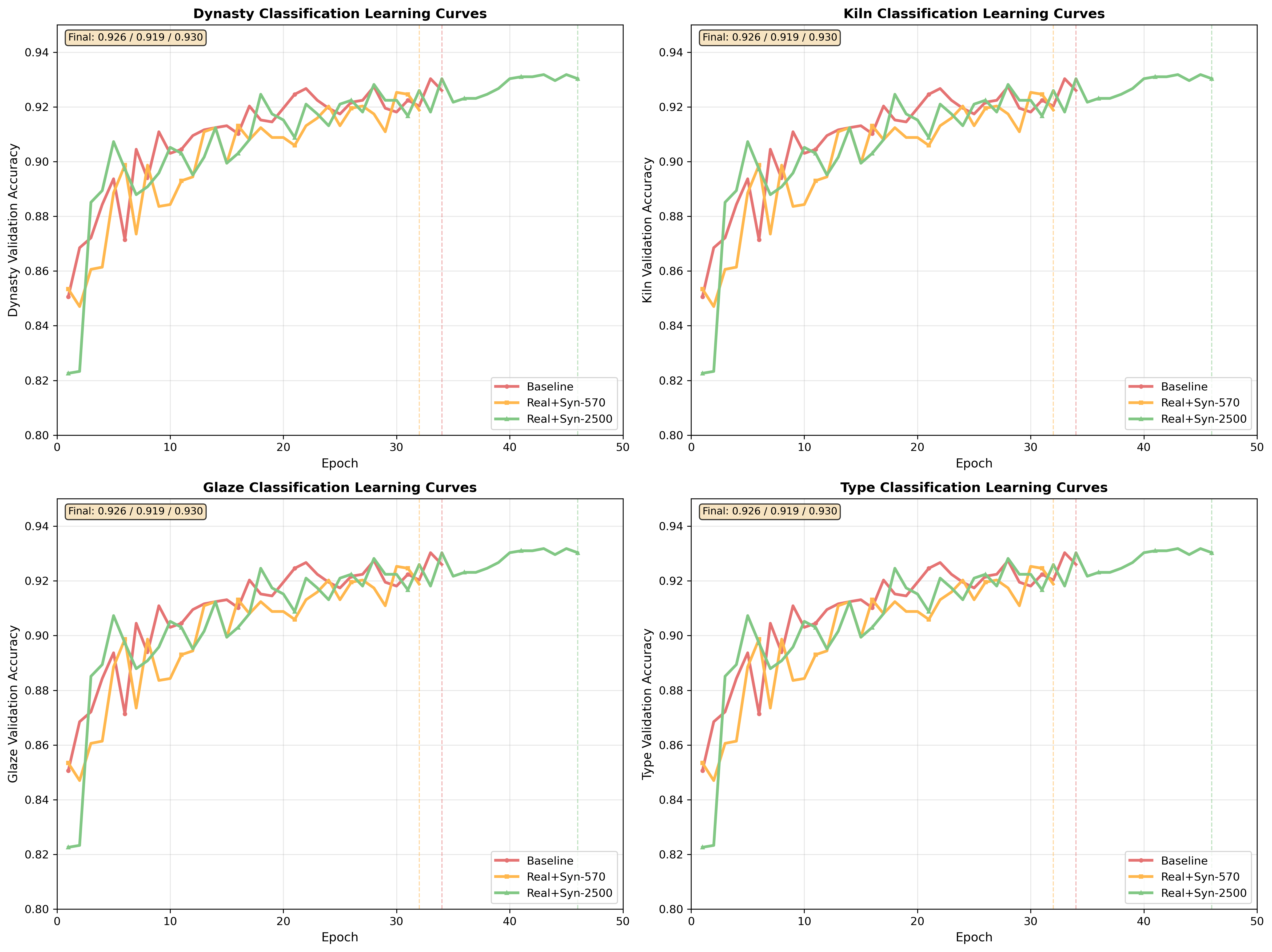}
    \caption{Task-wise Learning Dynamics and Convergence Patterns}\label{fig:/taskwise_dynamics}
\end{figure}

Figure \ref{fig:/taskwise_dynamics} presents task-wise learning curves, revealing heterogeneous effects. For \textbf{Dynasty}, all models converge similarly, but \textit{Real+Syn-2500} achieves the highest final accuracy (0.930), with notable gains for the minority Yuan dynasty (+7\% over baseline). For \textbf{Kiln}, final performance is similar across models, though \textit{Real+Syn-2500} shows smoother improvement for rarer kilns. \textbf{Glaze} remains the most challenging task, with all models exhibiting lower accuracy ($\approx$0.926) and higher volatility, reflecting synthesis limitations for fine-grained surface textures. In contrast, \textbf{Type} benefits most from synthetic data, with \textit{Real+Syn-2500} maintaining a clear advantage after epoch~20 and rare vessel types gaining 2--3\% in accuracy.

\begin{figure}[htbp]
    \centering
    \includegraphics[width=0.9\textwidth]{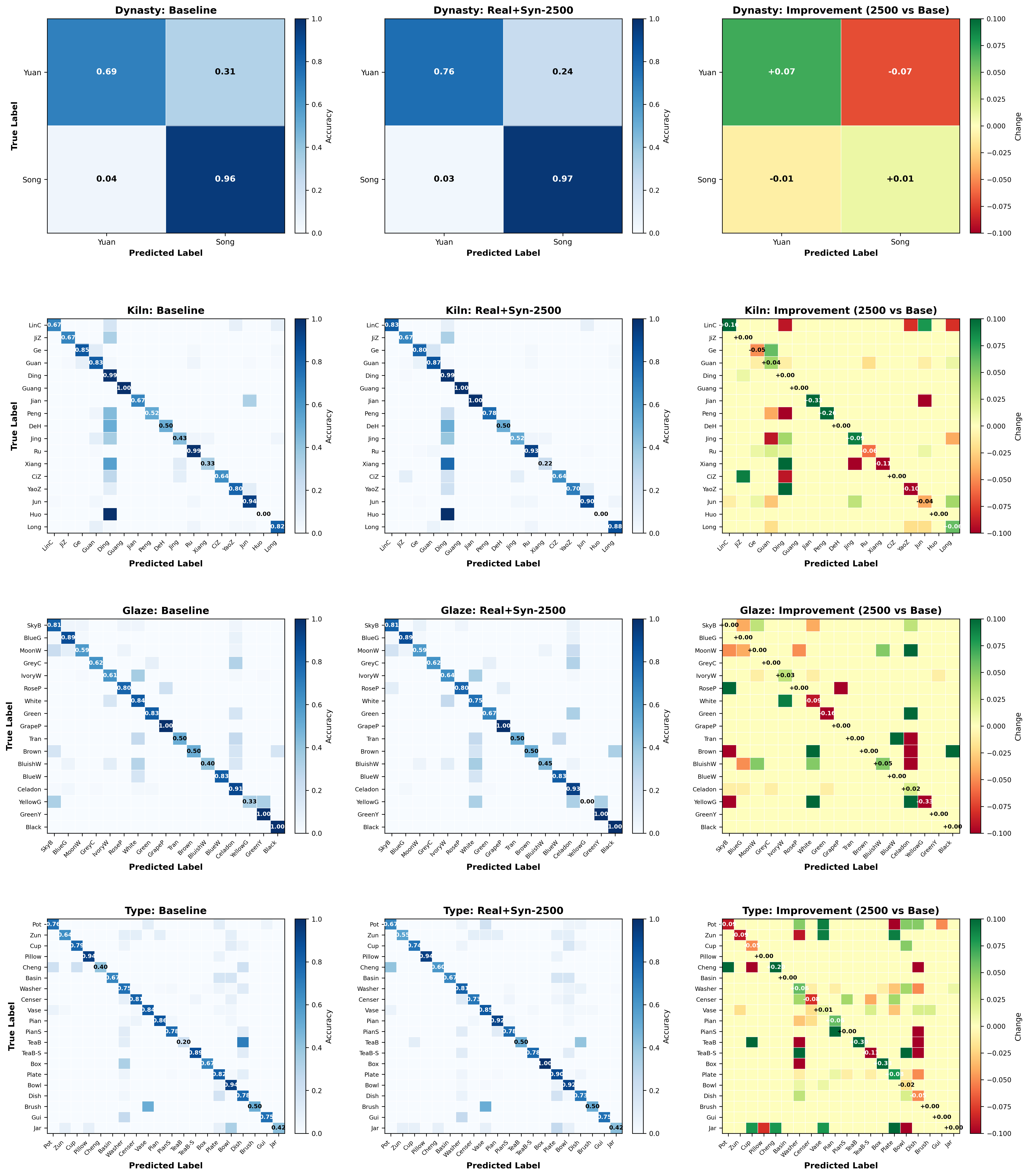}
    \caption{Confusion Matrix Analysis - Comparing Baseline and Real+Syn-2500 Across All Tasks}\label{fig:/confusion_matrix}
\end{figure}

The confusion matrix comparison in Figure \ref{fig:/confusion_matrix} further confirms these trends. Improvements are concentrated along the diagonal, indicating enhanced correct classification without introducing new confusion. Minority class gains are most evident in Dynasty classification (Yuan: 69\% $\rightarrow$ 76\%), while Kiln classification shows 2--3\% gains for rare kilns. Glaze classification displays mixed effects, with reduced accuracy for some visually similar glazes (e.g., BluishW vs. White). Type classification benefits broadly, especially for rare vessel forms(e.g., Cheng, Washer and Box).

In summary, larger synthetic proportions extend convergence but improve minority class performance, particularly in Dynasty and Type tasks. While training stability decreases with more synthetic data, the improvements are task-dependent and most pronounced when synthetic data comprises at least 10\% of the training set. These results indicate that targeted synthetic augmentation, when applied at sufficient scale, enhances classification fairness without substantially harming majority class accuracy, thereby addressing RQ1.3.

\subsection{Task-Specific Effectiveness}

Having established that synthetic augmentation requires at least 10\% dataset proportion for meaningful impact, this section examines the heterogeneous effects of the Real+Syn-2500 configuration across the four classification tasks. The analysis reveals fundamental relationships between task characteristics and synthetic data effectiveness that extend beyond simple class balance considerations. Table~\ref{tab:task_specific_performance} quantifies these task-specific impacts, illustrating both the potential benefits and the associated risks of integrating synthetic data. Dynasty and Type classification tasks exhibit consistent performance improvements with synthetic augmentation, likely due to the additional structural and morphological diversity introduced. In contrast, Glaze classification shows notable performance degradation, suggesting that the generation process does not sufficiently capture the surface texture nuances essential for glaze discrimination. To further understand these heterogeneous effects, we examine task-specific learning dynamics during training.”

\begin{table}[htbp]
\centering
\caption{Task-Specific Test Performance Impact}
\label{tab:task_specific_performance}
\renewcommand{\arraystretch}{1.2}
\begin{tabular}{|l|l|c|c|c|}
\hline
\textbf{Task} & \textbf{Metric} & \textbf{Traditional Only}  & \textbf{+Syn-2500} & \textbf{Best Change} \\
\hline
Dynasty & F1-macro & 0.8480  & 0.8808 & +3.9\% \\
        & Accuracy & 0.8964  & 0.9176 & +2.4\% \\
\hline
Kiln    & F1-macro & 0.7394  & 0.7615 & +3.0\% \\
        & Accuracy & 0.8689  & 0.8777 & +1.0\% \\
\hline
Glaze   & F1-macro & 0.7338  & 0.7079 & $-$3.5\% \\
        & Accuracy & 0.8115  & 0.7940 & $-$2.2\% \\
\hline
Type    & F1-macro & 0.7463  & 0.7791 & +4.4\% \\
        & Accuracy & 0.8240  & 0.8352 & +1.4\% \\
\hline
\end{tabular}
\end{table}

\begin{figure}[htbp]
    \centering
    \includegraphics[width=0.8\textwidth]{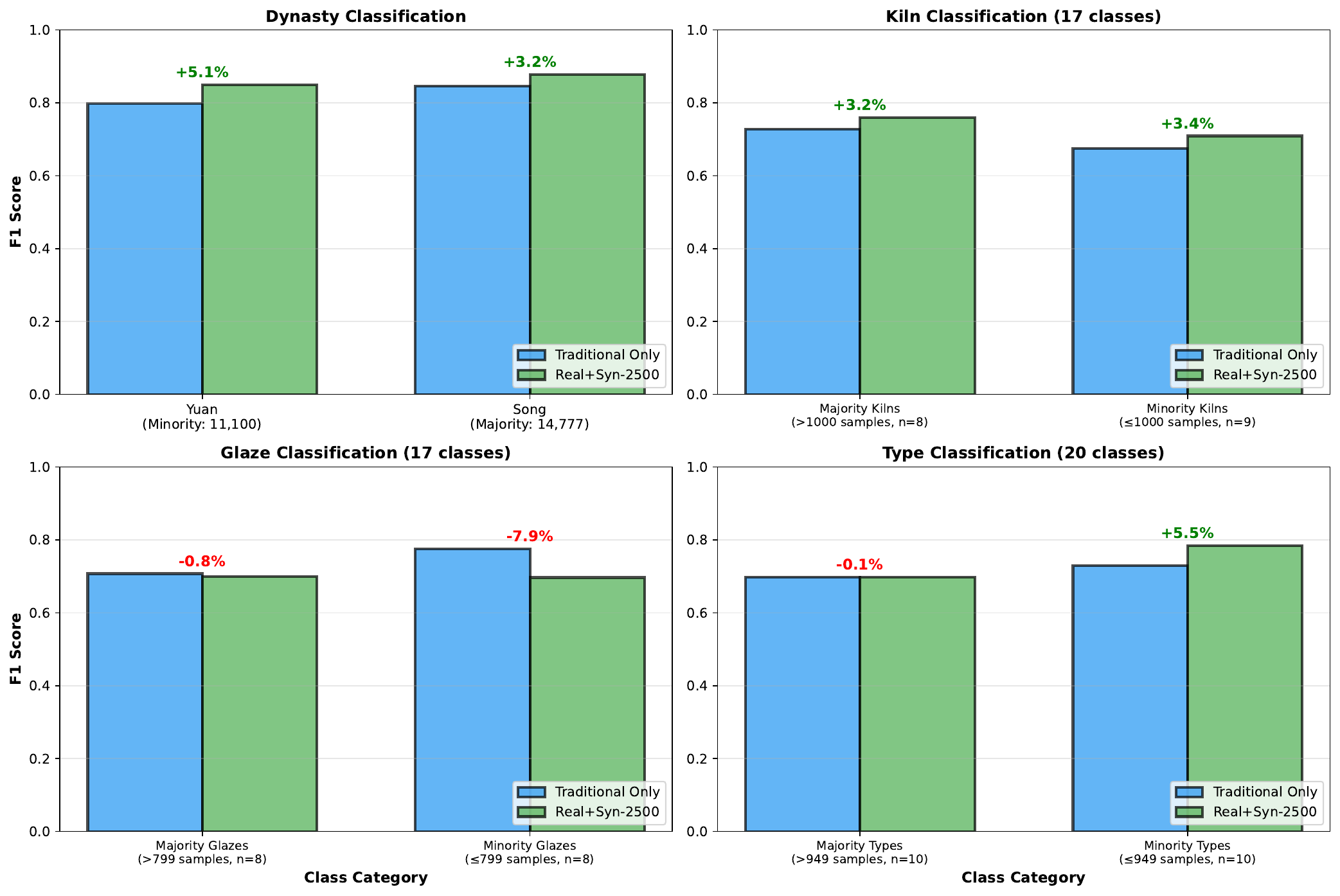}
    \caption{Minority vs. Majority Class Performance Gains with Synthetic Augmentation}
    \label{fig:/minority_majority}
\end{figure}

The most pronounced benefit of synthetic augmentation is observed in minority class performance, as shown in Figure \ref{fig:/minority_majority} and quantified by per-class F1 scores derived from the confusion matrices. Table \ref{tab:minority_majority_f1} presents the detailed breakdown of these differential impacts. These overall task-level differences are further reflected in minority–majority class behavior.

\begin{table}[htbp]
\centering
\caption{Minority vs. Majority Class F1 Improvements}
\label{tab:minority_majority_f1}
\setlength{\tabcolsep}{6pt}
\renewcommand{\arraystretch}{1.2}
\begin{tabular}{|p{1.2cm}|p{2.5cm}|p{1.5cm}|p{1.5cm}|p{1.5cm}|p{1.8cm}|}
\hline
\textbf{Task} & \textbf{Class Type} & \textbf{Samples} & \textbf{Baseline F1} & \textbf{+Syn-2500 F1} & \textbf{Improvement} \\
\hline
Dynasty & Yuan (minority)   & 11,100 & 0.794 & 0.844 & +5.1\% \\
        & Song (majority)   & 14,777 & 0.845 & 0.876 & +3.2\% \\
\hline
Kiln    & Majority ($>$1000) & 8 classes & 0.734 & 0.759 & +3.2\% \\
        & Minority ($\leq$1000) & 9 classes & 0.681 & 0.706 & +3.4\% \\
\hline
Glaze   & Majority ($>$799) & 8 classes & 0.708 & 0.702 & -0.8\% \\
        & Minority ($\leq$799) & 8 classes & 0.773 & 0.703 & -7.9\% \\
\hline
Type    & Majority ($>$949) & 10 classes & 0.703 & 0.702 & -0.1\% \\
        & Minority ($\leq$949) & 10 classes & 0.732 & 0.778 & +5.5\% \\
\hline
\end{tabular}
\end{table}

The results indicate task-dependent patterns in the impact of synthetic augmentation on class balance. In dynasty classification, the minority Yuan class benefits more substantially (+5.1\%) than the majority Song class (+3.2\%). Type classification exhibits an even greater disparity, with minority types gaining 5.5\% while majority types remain essentially unchanged (-0.1\%).

For kiln classification, improvements are similar for both majority and minority classes ($\sim$3.2--3.4\%), suggesting that regional porcelain features benefit uniformly from synthetic augmentation regardless of sample size. In contrast, glaze classification shows an inverse pattern, with minority glazes experiencing a larger performance decrease (-7.9\%) than majority glazes (-0.8\%), indicating that the generation process poorly captures rare glaze textures.

The heterogeneous effects across tasks appear to be closely associated with their reliance on different visual feature types. Table \ref{tab:visual_feature_impact} illustrates this relationship by mapping actual performance outcomes to the predominant visual features that each task relies on. The 4.4\% improvement in type classification---the highest among all tasks---suggests that morphological diversity is effectively captured by the LoRA-adapted diffusion model. Tasks relying on global structural features consistently benefit from synthetic data's capacity to generate diverse viewpoints while preserving overall proportions. In contrast, glaze classification's dependence on subtle surface textures, which are often smoothed during the diffusion process, likely accounts for its consistent performance degradation across both majority and minority classes.

\begin{table}[htbp]
\centering
\caption{Visual Feature Dependencies and Synthetic Impact}
\label{tab:visual_feature_impact}
\begin{tabular}{|p{3.0cm}|p{2.5cm}|p{2.5cm}|p{3.5cm}|}
\hline
\textbf{Visual Feature} & \textbf{Critical For} & \textbf{Actual Impact} & \textbf{Evidence from Data} \\
\hline
Global shape/profile & Dynasty, Type & +3.9\%, +4.4\% & Highest F1 improvements \\
\hline
Surface texture & Glaze & -3.5\% & Consistent degradation \\
\hline
Decorative patterns & Kiln & +3.0\% & Uniform improvement \\
\hline
Proportions/symmetry & Type & +4.4\% & Best overall performance \\
\hline
Regional markers & Kiln & +3.0\% & Stable across classes \\
\hline
\end{tabular}
\end{table}

Part of these differences may stem from the structured prompt engineering used during generation, which was derived from established archaeological documentation methods described in Subsection \ref{subsubsec: realimages}. Each prompt followed a hierarchical template informed by traditional morphological authentication principles and museum cataloging standards:

{\small
\begin{minipage}{\linewidth}
\ttfamily
Yuan dynasty, Jun kiln produced, Chinese porcelain, vase for display only, no spout or handle, decorative vessel, with moon white glaze with pale blue lighter than bluish green, thick opaque glaze
\end{minipage}
}

This structured approach incorporates five key archaeological descriptors. \textbf{Temporal attribution} (\enquote{Yuan dynasty}) was effectively captured in visual features, contributing to a +3.9\% F1-macro improvement in dynasty classification; period markers such as vessel proportions and rim profiles were successfully translated into visual characteristics. \textbf{Production site} (\enquote{Jun kiln produced}) was moderately captured, yielding a +3.0\% gain in kiln classification; distinctive traits such as the Jun kiln’s milky opacity were partially reproduced, although confusion persisted for kilns with similar glazing techniques. \textbf{Functional classification} (\enquote{rice bowl for daily meals} vs.\ \enquote{vase for display only}) achieved the largest improvement (+4.4\% in type classification); explicit functional descriptions aligned with morphological authentication principles, reflecting the traditional \enquote{form follows function} framework in scholarship. \textbf{Morphological descriptors} (\enquote{rounded shape, larger size}, \enquote{no spout or handle}) were strongly correlated with visual performance gains, consistent with the detailed form categories in Figure \ref{fig:formclass}; specific shape variations (e.g., \enquote{angled waist}, \enquote{lobed sides}) had clear visual manifestations. Finally, \textbf{Glaze characterization} (\enquote{moon white glaze with pale blue lighter than bluish green, thick opaque glaze}) showed the lowest performance change (--3.5\% in glaze classification); despite precise linguistic specification drawn from the glaze color categories in Figure \ref{fig:glazecolour}, fine-grained texture and color gradations proved difficult for the generative model to reproduce.

Prompt-guided generation effectively reduced several major confusions. For instance, the \textbf{TeaBowl$\rightarrow$Dish} error dropped by 30 percentage points when prompts explicitly distinguished \enquote{tea ceremony vessel} from \enquote{serving dish}, while the \textbf{Yuan$\rightarrow$Song} confusion decreased by 7 points with the inclusion of period-specific traits. In contrast, the \textbf{IvoryWhite$\rightarrow$White} confusion showed little change despite detailed color descriptions, underscoring the model’s persistent limitations in texture rendering. These results indicate that prompt optimization alone cannot overcome texture synthesis constraints. Future directions include \textbf{hierarchical prompting} that prioritizes reliably translated features such as shape and function; \textbf{task-specific templates} tailored to classification requirements; and the \textbf{integration of materials science terminology} (e.g., ``kaolin body,'' ``vitreous glaze'') to impose more precise generative constraints. Overall, while domain expertise enhances prompt design, successful augmentation requires recognizing the boundary between linguistically specifiable features and those that must be visually preserved.

Detailed analysis of the confusion matrices indicates that synthetic augmentation alters the distribution of classification errors in task-specific ways (Talbe \ref{tab:confusion_changes_syn2500}). The largest improvement is observed in type classification, where TeaBowl$\rightarrow$Dish confusion decreases by 30 percentage points, suggesting that synthetic data enhances shape discrimination through increased viewpoint diversity. In dynasty classification, Yuan$\rightarrow$Song confusion decreases by 7 points, indicating consistent gains. In contrast, glaze classification shows bidirectional degradation, with White$\rightarrow$IvoryWhite confusion increasing by 9 points, implying that synthetic generation may obscure subtle surface differences rather than resolve them.

\begin{table}[htbp]
\centering
\caption{Key Confusion Pattern Changes with Syn-2500}
\label{tab:confusion_changes_syn2500}
\begin{tabular}{|p{1cm}|p{2cm}|p{1.2 cm}|p{1.5cm}|p{1.2cm}|p{3cm}|}
\hline
\textbf{Task} & \textbf{Confusion Pair} & \textbf{Baseline} & \textbf{+Syn-2500} & \textbf{Change} & \textbf{Interpretation} \\
\hline
Dynasty & Yuan$\rightarrow$Song & 31\% & 24\% & $-7$\% & Better period discrimination \\
\hline
Kiln & LinC$\rightarrow$Ding & 17\% & 8\% & $-9$\% & Improved regional distinction \\
     & Jian$\rightarrow$Jian & 67\% & 100\% & $+33$\% & Better rare kiln recognition \\
\hline
Glaze & IvoryW$\rightarrow$White & 35\% & 34\% & $-1$\% & Minimal improvement \\
      & White$\rightarrow$IvoryW & 16\% & 25\% & $+9$\% & Increased reverse confusion \\
\hline
Type  & TeaB$\rightarrow$Dish & 70\% & 40\% & $-30$\% & Major shape clarification \\
      & Pot$\rightarrow$Vase & 10\% & 19\% & $+9$\% & New confusion introduced \\
\hline
\end{tabular}
\end{table}

Taken together, these observations converge on a consistent conclusion. The analysis confirms that synthetic augmentation is beneficial only when generative fidelity aligns with task requirements. Morphological and stylistic features (Dynasty +3.9\%, Type +4.4\%, Kiln +3.0\%) consistently improved, reflecting the model’s ability to capture shape and period markers. In contrast, glaze classification declined (--3.5\%), with minority glazes suffering the largest drop (--7.9\%), as subtle surface textures were poorly rendered. Error reductions such as TeaBowl$\rightarrow$Dish (–30.0\%) highlight augmentation’s value for morphology, while increased White$\rightarrow$IvoryWhite confusion illustrates its limits for texture. These results demonstrate that synthetic augmentation is effective for structure-based tasks but counterproductive for texture-dependent classification, underscoring the need for selective rather than universal application.

\section{ Discussion and Conclusion}
\label{sec:conclusion}

\subsection{Summary of Key Findings}

This paper presented a comprehensive investigation into the integration of LoRA-enhanced synthetic data generation with traditional augmentation strategies for archaeological porcelain classification. Comparing a baseline of 25,877 traditionally augmented images against configurations with 570 (5\%) and 2,500 (10\%) synthetic images, we established both the potential and limitations of synthetic augmentation in this domain. Overall, the Real+Syn-2500 configuration achieved a marginal 1.5\% gain in F1-macro (0.767$\rightarrow$0.782), yet the effect was highly task-dependent: dynasty (+3.9\%), type (+4.4\%), and kiln (+3.0\%) benefited consistently, while glaze declined by 3.5\%. This divergence highlights that synthetic augmentation is effective primarily for structure- and morphology-based tasks but counterproductive for texture-dependent classification. A clear threshold effect was observed: the Syn-570 configuration produced negligible or even negative impact, whereas at least 8--9\% synthetic data was required for measurable improvements, indicating that synthetic samples must form a sufficient proportion of training data to reshape the learned distribution. Per-class analysis further revealed that minority categories gained most, with Yuan dynasty (+5.1\%) and minority vessel types (+5.5\%) improving substantially, while minority glazes suffered a sharp decline (--7.9\%), underscoring the difficulty of synthesizing rare texture patterns. Rigorous quality control was essential, with 10\% of generated images rejected for historical or semantic inconsistencies. Finally, structured prompt engineering grounded in archaeological documentation proved effective for morphological descriptors (e.g., “rounded shape”), which reliably enhanced type classification, but failed to capture fine-grained glaze characteristics (e.g., “milky opacity”), confirming that linguistic specification cannot overcome fundamental limitations in texture synthesis. Together, these findings demonstrate that synthetic augmentation is a selective rather than universal tool: when aligned with task-relevant visual features, it improves representation and classification, but its benefits are constrained by the representational boundaries of current generative models.

\subsection{Theoretical and Practical Implications}

This work advances the understanding of synthetic-real distribution relationships in domain-specific computer vision tasks. The observed task-dependent performance variations provide empirical evidence that the effectiveness of synthetic data depends not on dataset size or class balance alone, but on the alignment between what generative models can faithfully reproduce and the discriminative features required by each task. The identified threshold at 10\% synthetic proportion suggests a phase transition in which synthetic patterns attain sufficient representation to influence feature learning, contributing to theories of mixed training dynamics in neural networks.

The differential impact across visual features, benefiting shape-based tasks while hindering texture-based tasks, supports the hypothesis that current diffusion models have an inherent smoothing bias, preserving global structure but degrading fine-grained surface details. This finding implies that synthetic augmentation strategies should be tailored to the specific visual features critical for each application domain.

This work provides empirical evidence for the limits of prompt-guided generation in specialized domains, showing that translation success decreases as feature abstractness increases. Despite the use of archaeologically grounded terminology from museum catalogs and scholarly documentation, such as "gold wire and iron thread" for Ge kilns or "fish scale crackles" for Ru kilns, performance varied by feature type: concrete morphological descriptors achieved high fidelity (+4.4\% in type classification), period markers showed moderate success (+3.9\% in dynasty classification), and surface texture descriptions consistently failed (--3.5\% in glaze classification). These findings contribute to understanding the semantic gap between expert linguistic descriptions and their visual realization in generative models.

The observed hierarchy of translation success, morphology $>$ period style $>$ regional characteristics $>$ surface texture, indicates that generative models are more effective at reproducing structural features than material properties. This finding clarifies the practical boundaries of synthetic data use in archaeological porcelain research and suggests that, while domain expertise is essential for prompt engineering, it cannot compensate for architectural limitations in texture synthesis.

These theoretical insights directly inform practical guidelines for the deployment of synthetic augmentation in archaeological porcelain research. Based on the empirical findings, we recommend a decision framework for the use of synthetic augmentation in archaeological porcelain classification. Synthetic augmentation should be applied when classification tasks rely primarily on morphological or structural features, when minority classes contain fewer than 100 samples after traditional augmentation, and when computational resources permit 3-4 hours of LoRA training. In such cases, even marginal improvements of 3--4\% can justify the additional computational cost. Conversely, synthetic augmentation should be avoided when texture fidelity is critical (e.g., glaze classification), when datasets already exceed 100 samples per class, when available computational resources are restricted to under 1 hour, or when quality control for generated data cannot be ensured. Effective prompt engineering should emphasize concrete morphological descriptors (e.g., "rounded shape", "larger size") and functional categories (e.g., "rice bowl for daily meals" vs.\ "vase for display only"), while using established archaeological terminology for dynasty and kiln attribution. However, detailed glaze descriptions such as "thick opaque glaze" cannot compensate for current generative model limitations in texture synthesis. Finally, the optimal configuration targets an 8--9\% synthetic proportion, adopts task-specific data pipelines in multi-task settings, and prioritizes augmentation of minority classes with structural rather than textural features.

\subsection{Limitations and Future Work}

The primary limitation of the current approach is the unresolved issue of texture degradation, with glaze classification consistently exhibiting performance losses. The inherent smoothing in the diffusion model's denoising process conflicts with the preservation of fine surface details. Despite sophisticated prompt engineering grounded in archaeological expertise, the language-to-visual translation remains constrained by the model's texture synthesis capabilities. Even precise terminology, such as "purple rim and iron foot" (Guan kilns) or "rabbit hair striation" (Jian kilns), fails to produce authentic surface characteristics.

Several promising avenues for further investigation emerge from these findings. First, task-adaptive generation could be developed by training separate LoRA adaptations optimized for distinct visual features, such as one specialized in shape and structure and another focused on texture and surface, to address the current trade-off between tasks. Second, multi-modal prompt enhancement may prove valuable by combining textual prompts with texture patches or technical specifications (e.g., surface roughness measurements or spectral reflectance data), thereby bridging the gap between linguistic descriptions and visual realizations. Third, the development of archaeological language models trained on museum catalogs, excavation reports, and scholarly literature could enable more accurate interpretation of domain-specific terminology, capturing subtle associations between descriptors such as \textit{celadon} and their visual traits. Fourth, texture-preserving architectures like VAE or dedicated texture synthesis networks should be explored to maintain fine-grained surface details that are critical for glaze classification. Finally, dynamic proportion optimization represents another promising direction, where adaptive algorithms determine the optimal proportion of synthetic data for each task based on validation performance, potentially leveraging reinforcement learning to balance multi-objective performance.

As generative models advance, particularly in texture synthesis and fine-detail preservation, the role of synthetic data in archaeological porcelain applications is likely to expand. Such expansion, however, should be guided by task-specific evaluation, domain expertise, and a clear understanding of the translation gap between linguistic description and visual realization. The contribution of this work lies in two areas: first, demonstrating the feasibility of integrating synthetic augmentation in archaeological porcelain classification; and second, establishing quantitative frameworks to determine both when synthetic augmentation delivers value in specialized domains and when linguistic specification reaches its limits in guiding visual generation.



\end{document}